\documentclass[sigconf]{acmart}
\usepackage{makecell}
\usepackage{multirow}
\usepackage{graphicx}
\usepackage{subfigure}
\usepackage[normalem]{ulem}
\usepackage{url}
\usepackage{enumerate}
\usepackage{booktabs}
\usepackage{multirow}
\usepackage{xspace}

\usepackage{amsmath}
\usepackage{color}
\usepackage{threeparttable}
\usepackage{ulem}
\usepackage{amssymb}

\usepackage{float}
\usepackage[ruled,linesnumbered]{algorithm2e}
\usepackage{algorithm2e, setspace, algpseudocode}

\def\bb#1{\mathbb{#1}}

\def\cal#1{\mathcal{#1}}

\newcommand{\eg}{\textit{e.g.,}\xspace}
\newcommand{\ie}{\textit{i.e.,}\xspace}

\newtheorem{definition}{Definition}

\AtBeginDocument{%
  \providecommand\BibTeX{{%
    \normalfont B\kern-0.5em{\scshape i\kern-0.25em b}\kern-0.8em\TeX}}}

\copyrightyear{2022} 
\acmYear{2022} 
\setcopyright{rightsretained} 

\acmConference[KDD '22]{Proceedings of the 28th ACM SIGKDD Conference on Knowledge Discovery and Data Mining}{August 14--18, 2022}{Washington, DC, USA}
\acmBooktitle{Proceedings of the 28th ACM SIGKDD Conference on Knowledge Discovery and Data Mining (KDD '22), August 14--18, 2022, Washington, DC, USA}
\acmDOI{10.1145/3534678.3539396}
\acmISBN{978-1-4503-9385-0/22/08}

\usepackage{etoolbox}
\makeatletter
\patchcmd{\maketitle}{\@copyrightpermission}{
   \begin{minipage}{0.3\columnwidth}
     \href{https://creativecommons.org/licenses/by/4.0/}{\includegraphics[width=0.90\textwidth]{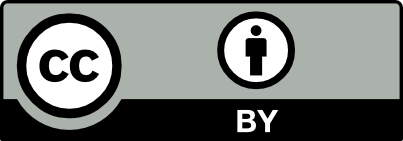}}
   \end{minipage}\hfill
   \begin{minipage}{0.7\columnwidth}
     \href{https://creativecommons.org/licenses/by/4.0/}{This work is licensed under a Creative Commons Attribution International 4.0 License.}
   \end{minipage}
  
   \vspace{5pt}
}{}{}

\makeatother

\begin{document}

\title{Pre-training Enhanced Spatial-temporal Graph Neural Network for Multivariate Time Series Forecasting}

% ================================== Author List 1 ================================== %
% \author{Zezhi Shao$^{\dagger}$, Zhao Zhang, Fei Wang$^{\ddagger}$, Yongjun Xu}
% \affiliation{%
%  \institution{Institute of Computing Technology, Chinese Academy of Sciences}
%  \city{Beijing}
%  \country{China}}
% \email{{shaozezhi19b, zhaozhang2021, wangfei, xyj}@ict.ac.cn}
% \thanks{$\dagger$Zezhi Shao is also with University of Chinese Chinese Academy of Sciences.
% \\
% $\ddagger$Corresponding author.}

% ================================== Author List 2 ================================== %
\author{Zezhi Shao}
\affiliation{
\institution{Institute of Computing Technology, \\Chinese Academy of Sciences}
\institution{University of Chinese Academy of Sciences}
\country{}
}
\email{shaozezhi19b@ict.ac.cn}

\author{Zhao Zhang}
\affiliation{
\institution{Institute of Computing Technology, \\Chinese Academy of Sciences}
\country{}
}
\email{zhangzhao2021@ict.ac.cn}

\author{Fei Wang}
\affiliation{
\institution{Institute of Computing Technology, \\Chinese Academy of Sciences}
\country{}
}
\email{wangfei@ict.ac.cn}
\authornote{Corresponding author.}

\author{Yongjun Xu}
\affiliation{
\institution{Institute of Computing Technology, \\Chinese Academy of Sciences}
\country{}
}
\email{xyj@ict.ac.cn}

\renewcommand{\shortauthors}{Zezhi Shao et al.}
\renewcommand{\authors}{Zezhi Shao, Zhao Zhang, Fei Wang, Yongjun Xu}

\begin{abstract}  
Multivariate Time Series~(MTS) forecasting plays a vital role in a wide range of applications.
Recently, Spatial-Temporal Graph Neural Networks~(STGNNs) have become increasingly popular MTS forecasting methods.
STGNNs jointly model the spatial and temporal patterns of MTS through graph neural networks and sequential models, significantly improving the prediction accuracy.
But limited by model complexity, most STGNNs only consider short-term historical MTS data, such as data over the past one hour.
However, the patterns of time series and the dependencies between them~(\ie the temporal and spatial patterns) need to be analyzed based on long-term historical MTS data.
To address this issue, we propose a novel framework, in which
\underline{S}\underline{T}GNN is \underline{E}nhanced by a scalable time series \underline{P}re-training model~(STEP).
Specifically, we design a pre-training model to efficiently learn temporal patterns from very long-term history time series~(\eg the past two weeks) and generate segment-level representations.
These representations provide contextual information for short-term time series input to STGNNs and facilitate modeling dependencies between time series.
Experiments on three public real-world datasets demonstrate that our framework is capable of significantly enhancing downstream STGNNs, and our pre-training model aptly captures temporal patterns.
\end{abstract}




\maketitle
\section{Introduction}
\label{sec_intro}

Multivariate time series data is ubiquitous in our lives, from transportation and energy to economics.
It contains time series from multiple interlinked variables.
Predicting future trends based on historical observations is of great value in helping to make better decisions.
% For example, traffic forecasting plays an important role in intelligent transportation systems since it is one of the foundations of traffic scheduling optimization.
Thus, multivariate time series forecasting has remained an enduring research topic in both academia and industry for decades.

\begin{figure}[t]
  \centering
  \setlength{\abovecaptionskip}{0.2cm}
  \setlength{\belowcaptionskip}{-0.4cm}
  \includegraphics[width=1\linewidth]{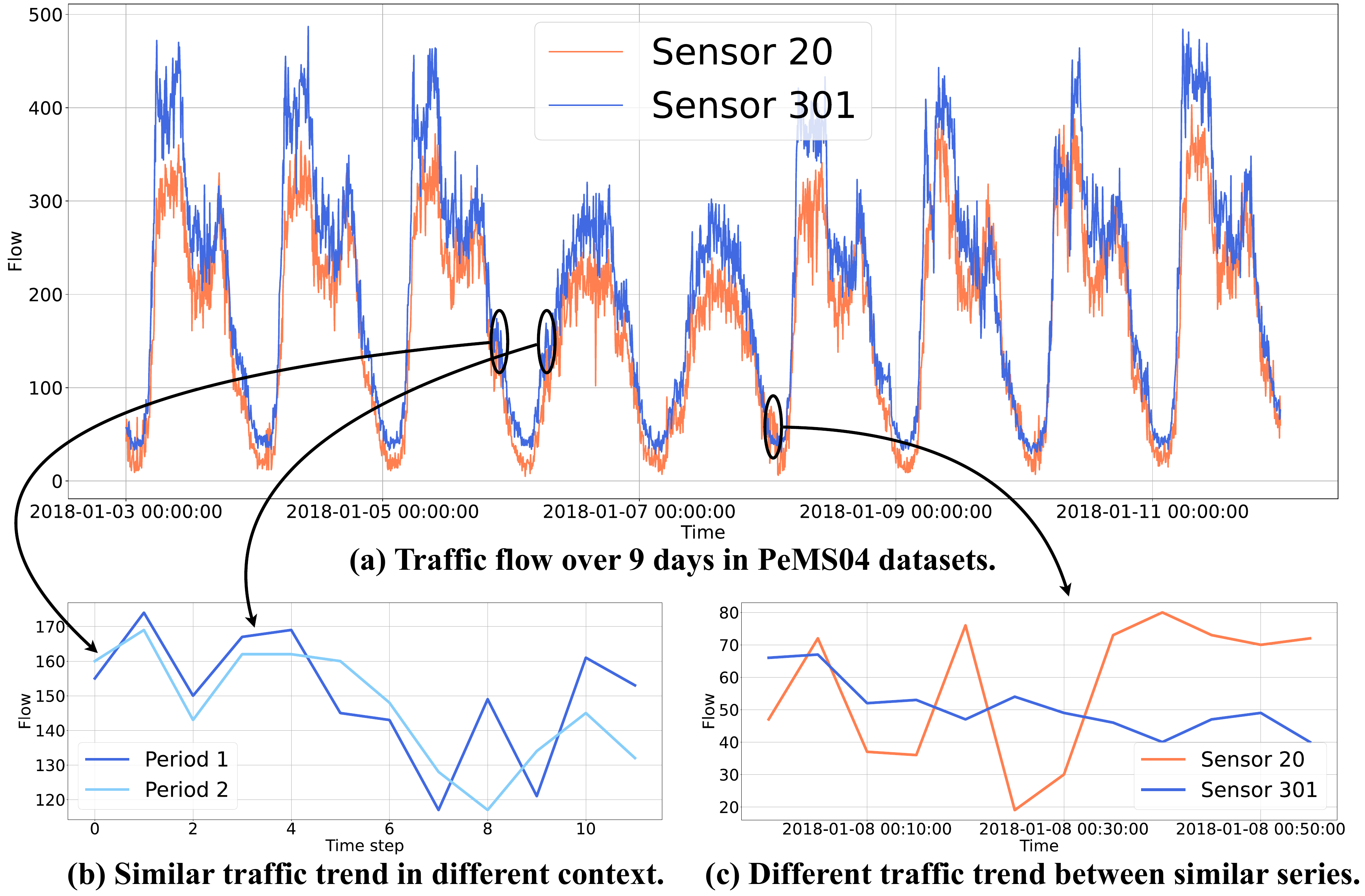}
  \caption{
%   \small
{
\fontsize{8.7pt}{\baselineskip}\selectfont 
  Examples of traffic flow multivariate time series data.
  (a) The two time series exhibit complex temporal patterns and strong spatial correlations.
  (b) Similar traffic trends within small windows in different contexts.
  (c) Different traffic trends within a small window between two similar series.
  }}
  \label{Intro}
\end{figure}

Indeed, multivariate time series can be generally formalized as spatial-temporal graph data~\cite{GWNet}.
On the one hand, multivariate time series have complex temporal patterns, \eg multiple periodicities.
On the other hand, different time series can affect other's evolutionary processes because of the underlying interdependencies between variables, which is non-Euclidean and is reasonably modeled by the graph structure.
To illustrate, we take the traffic flow system as an example, where each sensor corresponds to a variable.
Figure \ref{Intro}(a) depicts the traffic flow time series generated from two sensors deployed on the road network.
Apparently, there are two repeating temporal patterns, \ie daily and weekly periodicities.
The morning/evening peaks occur every day, while the weekdays and weekends exhibit different patterns.
Furthermore, the two time series share very similar trends because the selected sensors 20 and 301 are closely connected in the traffic network.
Consequently, accurate time series forecasting depends not only on the pattern of its temporal dimension but also on its interlinked time series.
Besides, it is worth noting that we made the above analysis on the basis of observing a sufficiently long time series.

To make accurate predictions, Spatial-Temporal Graph Neural Networks~(STGNNs) have attracted increasing attention recently.
STGNNs combine Graph Neural Networks~(GNNs)~\cite{2017GCN} and sequential models.
The former is used to deal with the dependencies between time series, and the latter is used to learn the temporal patterns.
% 
% For example, DCRNN~\cite{2017DCRNN} integrates diffusion graph convolution into Recurrent Neural Networks~(RNN).
% 
% The two most representative works are DCRNN~\cite{2017DCRNN} and Graph WaveNet~\cite{GWNet}.
% DCRNN integrates diffusion graph convolution into Recurrent Neural Networks~(RNN).
% Graph WaveNet uses diffusion graph convolution and dilated Convolution Neural Network~(CNN)~\cite{2016TCN} for achieving higher efficiency and performance.
Benefitting from jointly modeling the spatial and temporal patterns, STGNNs have achieved state-of-the-art performance.
In addition, an increasing number of recent works are further exploring the joint learning of graph structures and STGNNs since the dependency graph between time series, which is handcrafted by prior knowledge, is often biased and incorrect, even missing in many cases.
In short, spatial-temporal graph neural networks have made significant progress for multivariate time series forecasting in many real-world applications.
However, there is no free lunch.
More powerful models require more complex structures.
The computational complexity usually increases linearly or quadratically with the length of the input time series.
Further considering the number of time series (\eg hundreds), it is not easy for STGNNs to scale to very long-term historical time series.
In fact, most models use historical data in a small window to make predictions, \eg use the past twelve time steps (one hour) to predict the future twelve time steps~\cite{2020MTGNN, 2021GTS, 2020GMAN, 2017DCRNN, GWNet}.
The inability to explicitly learn from long-term information brings up some intuitive concerns.

Firstly, the STGNN model is blind to the context information beyond the window.
Considering that time series are usually noisy, it may be difficult for the model to distinguish short-term time series in different contexts.
For example, when observing data within two small windows of length twelve shown in Figure \ref{Intro}(b), we find that the two time series in different contexts are similar.
Therefore, it is difficult for models to make accurate predictions about their different future trends based on limited historical data.
Secondly, short-term information is unreliable for modeling the dependency graph, which is represented by the similarity (or correlation) between time series.
As shown in Figure \ref{Intro}(c), the two time series are not similar when we observe data within the small window, neither in number nor in trend.
On the contrary, long-term historical time series are beneficial for resisting noise, which facilitates obtaining more robust and accurate dependencies.
Although long-term historical information is beneficial, as mentioned above, it is expensive for the STGNNs to scale to very long-term historical time series directly.
% Furthermore, the optimization of the model can also become problematic with the increase of input sequence length.
Furthermore, the optimization of the model can also become problematic as the length of the input sequence increases.

% In order to solve the above problems, we hope to find another way and not directly design more complex STGNNs.
% Instead, a model is designed to efficiently extract very long-term historical information, and can effectively provide context information as bias for the raw short-term input of the STGNN models, and at the same time can well represent ultra-long sequences to solve the problem of graph structure learning.

% In order to solve the above problems, we hope to find another way and not directly design more complex STGNNs.
To address these challenges, we propose a novel framework, in which \underline{S}\underline{T}GNN is \underline{E}nhanced by a scalable time series \underline{P}re-training model~(STEP).
The pre-training model aims to efficiently learn the temporal patterns from very long-term historical time series and generate segment-level representations, which contain rich contextual information that is beneficial to address the first challenge.
In addition, the learned representations of these segments~(\ie the short-term time series) are able to incorporate the information from the whole long historical time series to calculate the correlation between time series, thus solving the second challenge, the problem of missing the dependency graph.
Specifically, we design an efficient unsupervised pre-training model for \underline{T}ime \underline{S}eries based on Trans\underline{Former} blocks~\cite{2017Transformer}~(TSFormer), which is trained through the masked autoencoding strategy~\cite{2021MAE}.
TSFormer efficiently captures information over very long-term historical data over weeks, and produces segment-level representations that correctly reflect complex patterns in time series.
Second, we design a graph structure learner based on the representation of TSFormer, which learns discrete dependency graph and utilizes the $k$NN graph computed based on the representation of TSFormer as a regularization to guide the joint training of graph structure and STGNN.
Notably, STEP is a general framework that can extend to almost arbitrary STGNNs.
In summary, the main contributions are the following:
\begin{itemize}
    \item We propose a novel framework for multivariate time series forecasting, where the STGNN is enhanced by a pre-training model.
    Specifically, the pre-training model generates segment-level representations that contain contextual information to improve the downstream models.
    \item We design an efficient unsupervised pre-training model for time series based on Transformer blocks and train it by the masked autoencoding strategy. 
    % The representations of TSFormer contain rich contextual information that can act as a bias.
    Furthermore, we design a graph structure learner for learning the dependency graph.
    \item Experimental results on three real-world datasets show that our method can significantly enhance the performance of downstream STGNNs, and our pre-training model aptly captures temporal patterns.
\end{itemize}

\section{Preliminaries}
We first define the concept of multivariate time series, the dependency graph.
Then, we define the forecasting problem addressed.

% Frequently used notations are summarized in Table 1.
\begin{definition}
\textbf{Multivariate Time Series.}
A multivariate time series has multiple time-dependent variable, such as observations from multiple sensors.
It can be denoted as a tensor $\mathcal{X}\in\mathbb{R}^{T\times N\times C}$, where $T$ is the number of time steps, $N$ is the number of variables, \eg the sensors, and $C$ is the number of channels.
We additionaly denote the data of time series $i$ as $\mathbf{S}^i\in\mathbb{R}^{T\times C}$.
\end{definition}

\begin{definition}
\textbf{Dependency Graph.}
Each variable depends not only on its past values but also on other variables. Such dependencies are captured by a dependency graph $\mathcal{G}=(V, E)$, where $V$ is the set of $|V|=N$ nodes, and each node corresponds to a variable, \eg a sensor. $E$ is the set of $|E|=M$ edges.
The graph can also be denoted as an adjacent matrix $\mathbf{A}\in\mathbb{R}^{N\times N}$.
\end{definition}

\begin{definition}
\textbf{Multivariate Time Series Forecasting.}
Given historical signals $\mathcal{X}\in\mathbb{R}^{T_h\times N\times C}$ from the past $T_h$ time steps, multivariate time series forecasting aims to predict the values $\mathcal{Y}\in\mathbb{R}^{T_f\times N\times C}$ of the $T_f$ nearest future time steps.
\end{definition}

\section{Model Architecture}
\begin{figure*}[t]
  \centering
  \setlength{\abovecaptionskip}{0.2cm}
  \setlength{\belowcaptionskip}{-0.2cm}
  \includegraphics[width=0.985\linewidth]{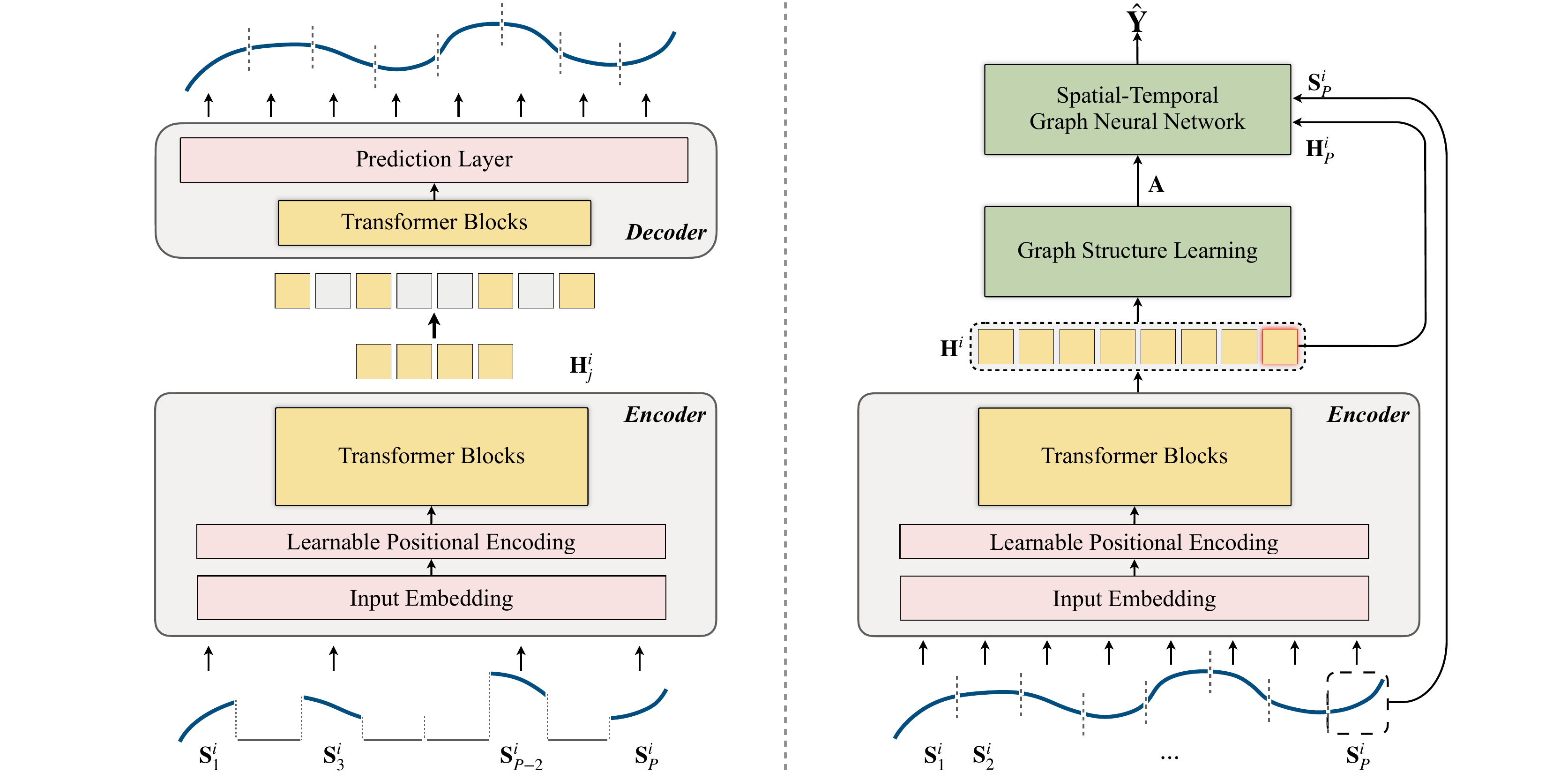}
  \caption{
  {\color{black}
  The overview of the proposed STEP framework.
  \underline{Left}: the pre-training stage. We split very long-term time series into segments and feed them into TSFormer, which is trained via the masked autoencoding strategy.
  \underline{Right}: the forecasting stage. We enhance the downstream STGNN based on the segment-level representations of the pre-trained TSFormer.}
  }
  \label{model}
\end{figure*}
As shown in Figure \ref{model}, STEP has two stages: the pre-training stage and the forecasting stage.
In the pre-training stage, we design a masked autoencoding model for \underline{T}ime \underline{S}eries based on Trans\underline{Former} blocks~(TSFormer) to efficiently learn temporal patterns.
TSFormer is capable of learning from the very long-term sequence and gives segment-level representations that contain rich context information.
In the forecasting stage, we use the pre-trained encoder to provide context information to enhance the downstream STGNN.
Furthermore, based on the representations of the pre-training model, we further design a discrete and sparse graph learner to deal with the cases that the pre-defined graph is missing.

\vspace{-0.1cm}
\subsection{The Pre-training Stage}
In this part, we aim at designing an efficient unsupervised pre-training model for time series.
While the pre-training model has made significant progress in natural language processing~\cite{2019BERT, 2020GPT, 2018ELMO}, progress in time series lags behind them.
First, we would like to discuss the difference between time series and natural language, which will motivate the design of TSFormer.
We attempt to distinguish them from the following two perspectives:

\textbf{(i) Time series information density is lower.} 
% Compared with natural language, data points in time series give less semantic information.
As human-generated signals, each data point in natural language~(\ie a word in a sentence) has rich semantics and is suitable as the data unit for model input.
On the contrary, isolated 
{\fontsize{8.8pt}{\baselineskip}\selectfont 
data points in time series} give less semantic information. 
Semantics only arise when we observe at least segment-level data, such as going up or down.
On the other hand, language models are usually trained by predicting only a few missing words per sentence.
However, masked values in time series can often be trivially predicted by simple interpolation~\cite{2021KDD_Transformer}, making the pre-training model only focuses on low-level information.
% , \ie, numerical information
To address this problem, a simple strategy that works well is to mask a very high portion of the model's input to encourage the learning of high-level semantics, motivated by recent development in computer vision~\cite{2021MAE}. 
% A simple strategy of masking a very high portion of the model’s input will work well to encourage learning of high-level semantics, motivated by recent development in computer vision~\cite{2021MAE}. 
% To encourage learning of high-level semantics, motivated by recent development in computer vision~\cite{2021MAE}, a simple strategy of masking a very high portion of the model's input will work well. 
It creates a challenging self-supervised task that forces the model to obtain holistic understanding of time series.

\textbf{(ii) Time series require longer sequences to learn the temporal patterns.}
In natural languages, sequences of hundreds of lengths have contained rich semantic information.
Thus, language pre-training models usually cut or pad the input sequence to hundreds~\cite{2018ELMO, 2020GPT, 2019BERT}.
However, although time series have relatively more straightforward semantics than natural languages, they require longer sequences to learn it. For example, traffic system records data every five seconds, and if we want to learn the weekly periodicity, we need at least consecutive 2016 time slices. 
Although sampling at a lower frequency is a possible solution, it inevitably loses information.
Fortunately, although longer time series will increase the model complexity, we can alleviate it by stacking fewer Transformer blocks and fix model parameters during the forecasting stage to reduce computational and memory overhead.
% \textbf{(iii) Frequency domain is important for obtain significant features.}
% 不同的语义在时间序列中的时域中可能非常相似。
% 例如，序列[70, 68, 66, 64] 和 [64, 66, 68, 70]在时域中相似度~(例如cosine相似度）较高，但他们代表上升和下降两种语义。
% % 频率领域更容易获取到可区分的特征
% There is little noise in language and images, which basically does affect the discrimination of patterns.
% In non-adversarial situations, noise generally only appears in individual places (such as typos in natural language) or low-amplitude global noise (such as noise in images).
% Due to the noise of the sensor and the stability of the system, the time series generally has global noise with high signal-to-noise ratio.

% 基于以上分析以及CV预训练模型最新工作的启发，我们。。。
Motivated by the above analyses and recent computer vision models~\cite{2020ViT}, especially Masked AutoEncoder~(MAE)~\cite{2021MAE}, we propose a masked autoencoding model for time series based on Transformer blocks~(\ie TSFormer).
TSFormer reconstructs the original signal based on the given partial observed signals.
We use an asymmetric design to largely reduce computation: the encoder operates on only partially visible signals, and the decoder uses a lightweight network on the full signals.
The model is shown in Figure \ref{model}(left), and we will introduce each component in detail next.

\noindent \textbf{Masking.} 
{\color{black}
We divide the input sequence $\mathbf{S}^{i}$ from node $i$ into $P$ non-overlapping patches of length $L$~(Input sequences are obtained over the original time series through a sliding window of length $P*L$).
The $j$th patch can be denoted as $\mathbf{S}_j^i\in\mathbb{R}^{LC}$, where $C$ is the input channel.}
We assume $L$ is the commonly used length of input time series of STGNNs.
We randomly mask a subset of patches with masking ratio $r$ set to a high number of 75$\%$ to create a challenging self-supervised task.
Here, we emphasize that the strategy of using patches as input units serves multiple purposes.
Firstly, segments~(\ie patches) are more appropriate for explicitly providing semantics than separate points.
Secondly, it facilitates the use of downstream models, as downstream STGNNs take a single segment as input.
Last but not least, it significantly reduces the length of sequences input to the encoder, and the high masking ratio $r$ makes the encoder more efficient during the pre-training stage.

% First, it significantly reduces the length of sequences input to the encoder.
% Second, it facilitates the use of downstream models since downstream STGNNs also take a sequence of length $L$ as input.
% Last but not least, the highly masking ratio $r$ makes the encoder very efficient during the pre-training stage.

\noindent \textbf{Encoder.} 
% Standard Transformer Blocks
% Embedding Layer
% Only operates on observed pathces
% Learnable positional encoding~(Different from MAE)
    % Reason 1: better performance
    % Reason 2: reflects the periodicity, which is demonstrated in Section X.
% 4Layer
% Our encoder is a series of Transformer blocks with a linear projection~(\ie the input embedding layer) and a positional embedding layer.
% The encoder only operates on unmasked patches.
% As the semantics of time series are more straightforward, we use four layers of  Transformer blocks, far less than the depth of Transformer-based models in computer vision~\cite{2020ViT, 2021MAE} and natural languages~\cite{2019BERT, 2020GPT}. 
% Notably, the positional embeddings are added to all patches, although the mask tokens are not used in the encoder. 
% Moreover, unlike the deterministic, sinusoidal embeddings used in MAE~\cite{2021MAE}, we use learnable positional embeddings. 
% On the one hand, in this work, learnable embeddings significantly outperform sinusoidal ones for all datasets. 
% On the other hand, we observed that learned positional embeddings is crucial in learning time series' periodic features, which will be demonstrated in Section \ref{sec_inspecting}.
 Our encoder is a series of Transformer blocks~\cite{2017Transformer} with an input embedding layer and a positional encoding layer. 
The encoder only operates on unmasked patches.
As the semantics of time series are more straightforward than languages, we use four layers of  Transformer blocks, far less than the depth of Transformer-based models in computer vision~\cite{2020ViT, 2021MAE} and natural languages~\cite{2019BERT, 2020GPT}. 
Specifically, the input embedding layer is a linear projection to transform the unmasked patches into latent space:
\begin{equation}
\setlength\abovedisplayskip{0.2cm}
\setlength\belowdisplayskip{0.2cm}
    \mathbf{U}_j^i = \mathbf{W}\cdot\mathbf{S}^i_j + \mathbf{b},
\end{equation}
where $\mathbf{W}\in\mathbb{R}^{d\times(LC)}$ and $\mathbf{b}\in\mathbb{R}^d$ are learnable parameters, $\mathbf{U}^i_j\in\mathbb{R}^{d}$ are the model input vectors, and $d$ is the hidden dimension.
For masked patches, we use a shared learnable mask token to indicate the presence of a missing patch to be predicted.
Next, the positional encoding layer is used to add sequential information.
Notably, the positional encoding operates on all patches, although the mask tokens are not used in the encoder. 
Moreover, unlike the deterministic, sinusoidal embeddings used in MAE~\cite{2021MAE}, we use learnable positional embeddings. 
On the one hand, in this work, learnable embeddings significantly outperform sinusoidal ones for all datasets. 
On the other hand, we observe that learned positional embeddings are crucial in learning time series' periodic features, which will be demonstrated in Section \ref{sec_inspecting}.
Finally, we obtain the latent representations $\mathbf{H}_j^i\in\mathbf{R}^d$ through Transformer blocks for all unmasked patches $j$.

\noindent \textbf{Decoder.} 
The decoder is also a series of Transformer blocks that reconstruct the latent representations back to a lower semantic level, \ie numerical information. 
The decoder operates on the full set of patches, including the mask tokens.
% We use a shared learnable mask token to indicate the presence of a missing patch to be predicted. 
Unlike MAE~\cite{2021MAE}, we no longer add positional embeddings here since all patches already have positional information added in the encoder. 
Notably, the decoder is only used during the pre-training stage to perform the sequence reconstruction task, and can be designed independently of the encoder.
We use only a single layer of Transformer block for balancing efficiency and effectiveness.
% Finally, we apply a Multi-Layer Perception~(MLP) to make predictions whose number of output dimensions equals the length of each patch, \ie $\text{L}$.
{\color{black}
Finally, we apply Multi-Layer Perceptions~(MLPs) to make predictions whose number of output dimensions equals the length of each patch.
Specifically, given the latent representation $\mathbf{H}^i_j\in\mathbb{R}^d$ of patch $j$, the decoder gives the reconstructed sequence $\hat{\mathbf{S}}^i_j\in\mathbb{R}^{LC}$.}

\noindent \textbf{Reconstruction target.}
Our loss function compute mean absolute error between the original sequence $\mathbf{S}^i_j$ and reconstructed sequence $\hat{\mathbf{S}}^i_j$.
Kindly note that we only compute loss over the masked patches, which is in line with other pre-training models~\cite{2021MAE, 2019BERT}. Moreover, all these operations are computed in parallel for all time series $i$.
% After the projection layer, we will get a reconstructed time series.
% Our model is trained by predicting sequence of each masked patches.
% We use mean absolute error as the distance the reconstructed and original sequence.
% We adopt mean absolute error as the loss function for the reconstructed and original sequence.
% Kindly note that the loss is only computed on the masked patches.
% Note that the loss is only computed on the masked patches.

In summary, TSFormer is efficient thanks to the high masking ratio and fewer Transformer blocks.
TSFormer is capable of learning from the very long-term sequence (\eg weeks) and can be trained on a single GPU.
The encoder generates representations for the input patches~(segments).
% and their concatenation can represent the entire input sequence. 
Furthermore, another noteworthy difference from MAE~\cite{2021MAE} is that we pay more attention to the representations of the patches. 
On the one hand, we can use the representations to verify periodic patterns in the data, which will be demonstrated in Section \ref{sec_inspecting}. 
More importantly, they can conveniently act as contextual information for short-term input of downstream STGNNs, which will be introduced in the next.
% More details and formulas can be found in Appendix \ref{appendix_tsformer}.
% downstream models since the STGNNs take the last patch as input.

\subsection{The Forecasting Stage}
% 预测阶段：我们使用预训练模型增强时空图神经网络。
% 输入输出：对于给定的时间片，STGNN通常选取较短的L长度的历史数据为输入。TSFormer将这个输入从L扩展到了P*L，产生了P个对应的表征H_i?。
% 我们使用这些表征建模缺失的依赖图，并增强下游时空图神经网络。
For a given time series $i$, TSFormer takes its historical signals $\mathbf{S}^i\in\mathbb{R}^{T_p\times C}$ of the past $T_p=L\times P$ time steps as input.
We divide it into $P$ non-overlapping patches of length $L$: $\mathbf{S}_1^i, \cdots, \mathbf{S}_P^i$, where $\mathbf{S}_j^i\in\mathbb{R}^{L\times C}$.
The pre-trained TSFormer encoder generates representations $\mathbf{H}_j^i\in\mathbb{R}^{d}$ for each $\mathbf{S}_j^i$, where $d$ is the dimension of hidden states.
Considering that the computational complexity usually increases linearly or quadratically with the length of the input time series, STGNNs can only take the latest, \ie the last patch $\mathbf{S}_P^i\in\mathbb{R}^{L\times C}$ for each time series $i$ as input.
For example, the most typical setting is $L=12$.
In the forecasting stage, we aim at enhancing the STGNNs based on the representations of the pre-trained TSFormer encoder.

% STGNN依赖一个图
\noindent\textbf{Graph structure learning.} Many STGNNs~\cite{2017DCRNN, GWNet,2020GMAN} depend on a pre-defined graph to indicate the relationship between nodes (\ie time series). 
However, such a graph is not available or is incomplete in many cases.
% 直观的想法
An intuitive idea is to train a matrix $\mathbf{A}\in\mathbb{R}^{N\times N}$, where $\mathbf{A}_{ij}\in[0,1]$ indicates the dependency between time series $i$ and $j$.
% 难点 learn the spatial dependencies in a supervised manner.
However, since the learning of graph structure and STGNNs are coupled compactly, and there is no supervised loss information for graph structure learning~\cite{2021REST}, optimizing such a contiguous matrix usually leads to a complex bilevel optimization problem~\cite{2019LDS}.
In addition, the dependency $\mathbf{A}_{ij}$ is usually measured by the similarity between time series, which is also a challenging task.

Fortunately, we can alleviate these problems based on the pre-trained TSFormer.
Motivated by recent works~\cite{2018NRI, 2019LDS, 2021GTS}, we aim to learn a discrete sparse graph, where $\mathbf{\Theta}_{ij}$ parameterizes the Bernoulli distribution from which the discrete dependency graph $\mathbf{A}$ is sampled.
% 做法
First, we introduce graph regularization to provide supervised information for graph optimization based on the representations of TSFormer.
Specifically, we denote $\mathbf{H}^i=\mathbf{H}^i_1\parallel\mathbf{H}^i_2...\mathbf{H}^i_{P-1}\parallel\mathbf{H}^i_{P}\in\mathbb{R}^{Pd}$ as the feature of time series $i$, where $\parallel$ means the concatenation operation.
Then we calculate a $k$NN graph $\mathbf{A}^a$ among all the nodes. We can control the sparsity of the learned graph by setting different $k$.
Benefiting from the ability of TSFormer, $\mathbf{A}^a$ can reflect the dependencies between nodes, which is helpful to guide the training of the graph structure.
Then, we compute $\mathbf{\Theta}_{ij}$ as follows:
\begin{equation}
    \begin{aligned}
    \mathbf{\Theta}_{ij}&=\text{FC}(\text{relu}(\text{FC}(\mathbf{Z}^i
    \parallel \mathbf{Z}^j)))\\
    \mathbf{Z}^i&= \text{relu}(\text{FC}(\mathbf{H}^i)) + \mathbf{G}^i,
    \end{aligned}
\end{equation}
where $\mathbf{\Theta}_{ij}\in \mathbb{R}^{2}$ is the unnormalized probability. The first dimension indicates the probability of positive, and the second dimension indicates  the probability of negative.
{\color{black}$\mathbf{G}^i$ is the global feature of time series $i$, which is obtained by a convolutional network $\mathbf{G}^i=\text{FC}(\text{vec}(\text{Conv}(\mathbf{S}^i_{train})))$, where
$\mathbf{S}^i_{train}\in \mathbb{R}^{L_{train}}$ is the entire sequence $i$ over training dataset, and $L_{train}$ is the length of the training dataset.}
$\mathbf{S}^i_{train}$ is static for all samples during training, helping to make the training process more robust and accurate.
The feature $\mathbf{H}^i$ is dynamic for different training samples to reflect the dynamics of dependency graphs~\cite{2021DGCRN}.
As such, we use the cross-entropy between $\mathbf{\Theta}$ and the $k$NN graph $A^a$ as graph structure regularization:
\begin{equation}
    \mathcal{L}_{graph} = \sum_{ij}-\mathbf{A}_{ij}^a\log\mathbf{\Theta}^{'}_{ij}-(1-\mathbf{A}_{ij}^{a})\log(1-\mathbf{\Theta}^{'}_{ij}),
\end{equation}
where $\mathbf{\Theta}^{'}_{ij}=\text{softmax}(\mathbf{\Theta}_{ij})\in\mathbb{R}$ is the normalized probability.

The last problem of discrete graph structure learning is that the sampling operation from $\mathbf{\Theta}$ to adjacent matrix $\mathbf{A}$ is not differentiable.
Hence, we apply the Gumbel-Softmax reparametrization trick proposed by \cite{Gumbel1, Gumbel2}:
\begin{equation}
    \mathbf{A}_{ij}=\text{softmax}((\mathbf{\Theta}_{ij}+\mathbf{g})/\tau),
\end{equation}
where $\mathbf{g}\in \mathbb{R}^2$ is a vector of i.i.d. samples drawn from a $\text{Gumbel(0,1)}$ distribution.
$\tau$ is the softmax temperature parameter.
The Gumbel-Softmax converges to one-hot samples~(\ie discrete) when $\tau \rightarrow 0$.

\noindent\textbf{Downstream spatial-temporal graph neural network.}
A normal downstream STGNN takes the last patch and the dependency graph as input, while the enhanced STGNN also considers the input patch's representation.
Since the TSFormer has strong power at extracting very long-term dependencies, the representation $\mathbf{H}^i_P$ contains rich context information.
STEP framework can extend to almost any STGNN, and we choose a representative method as our backend, the Graph WaveNet~\cite{GWNet}.
Graph WaveNet captures spatial-temporal dependencies efficiently and effectively by combining graph convolution with dilated casual convolution. 
It makes predictions based on its output latent hidden representations $\mathbf{H}_{gw}\in\mathbb{R}^{N\times d'}$ by a regression layer, which is a Multi-Layer Perception~(MLP).
{\color{black}For brevity, we omit its details, and interested readers can refer to the paper~\cite{GWNet}}.
Denoting the representations $\mathbf{H}^i_P$ of TSFormer for all node $i$ as $\mathbf{H}_P\in\mathbb{R}^{N\times d}$, we fuse the representations of Graph WaveNet and TSFormer by:
\begin{equation}
    \mathbf{H}_{final}=\text{SP}(\mathbf{H}_P) + \mathbf{H}_{gw},
    \label{fuse}
\end{equation}
where $\text{SP}(\cdot)$ is the semantic projector to transform the $\mathbf{H}^i_P$ to the semantic space of $\mathbf{H}_{gw}$. We implement it with a MLP.
Finally, we make predictions by the regression layer:  
$\hat{\mathcal{Y}}\in\bb{R}^{T_f\times N \times C}$. 
Given the ground truth $\cal{Y}\in\bb{R}^{T_f\times N \times C}$, we use mean absolute error as the regression loss:
\begin{equation}
    \cal{L}_{regression} = \cal{L}(\hat{\cal{Y}}, \cal{Y})=\frac{1}{T_fNC}
    \sum_{j=1}^{T_f}\sum_{i=1}^{N}\sum_{k=1}^{C}|\hat{\cal{Y}}_{ijk} - \cal{Y}_{ijk}|, 
    \label{loss}
\end{equation}
where $N$ is the number of nodes, $T_f$ is the number of forecasting steps, and $C$ is the dimensionality of the output.
The downstream STGNN and the graph structure is trained in an end-to-end manner:
\begin{equation}
    \mathcal{L} = \mathcal{L}_{regression} + \lambda \mathcal{L}_{graph}.
    \label{full_loss}
\end{equation}
We set the graph regularization term $\lambda$ gradually decay during the training process to go beyond the $k$NN graph.
Notably, the pre-trained TSFormer encoder is fixed in the forecasting stage to reduce computational and memory overhead.

\section{Experiments}
% In this section, we present experiments on three commonly used real-world datasets to demonstrate the effectiveness of STEP, including the proposed TSFormer and the enhanced STGNN.
% 我们首先介绍实验设置，包括数据集、Baseline和参数设定。
% 之后，我们对比D2STGNN和Baselines在两个数据集的上的性能。
% 然后，我们设计实验验证了Decoupled Framework的优越性。
% 最后，我们做了详尽的消融实验和参数实验，来验证我们方法的每个部件的作用。
% We first introduce the experimental settings, including datasets, baselines, metrics, and parameter settings.
% Then, we conduct experiments to compare the performance of STEP with other baselines.
% Furthermore, we design more experiments to demonstrate the ability of the TSFormer in learning temporal patterns.
% Finally, we design comprehensive ablation studies to evaluate the impact of important hyper-parameters and components.

% {\color{blue}Following previous works~\cite{2021GTS, 2020MTGNN}}, 
In this section, we present experiments on three real-world datasets to demonstrate the effectiveness of the proposed STEP and TSFormer.
Furthermore, we conduct comprehensive experiments to evaluate the impact of important hyper-parameters and components.
More experimental details, such as optimization settings and efficiency study, can be found in Appendix \ref{appendix_experiments}, \ref{appendix_efficiency}, and \ref{appendix_visualization}.
It is notable that we conduct pre-training for each dataset since these datasets are heterogeneous in terms of length of time series, physical nature, and temporal patterns. 
Our code can be found in this repository\footnote{\url{https://github.com/zezhishao/STEP}}.

\subsection{Experimental Setup}
\noindent\textbf{Datasets.}
Following previous works~\cite{2021GTS, 2020MTGNN, GWNet}, we conduct experiments on three commonly used multivariate time series datasets:
\begin{itemize}
    \item \textbf{METR-LA} is a traffic speed dataset collected from loop-detectors located on the LA County road network~\cite{METR-LA}. 
    It contains data of 207 selected sensors over a period of 4 months from Mar to Jun in 2012~\cite{2017DCRNN}. 
    The traffic information is recorded at the rate of every 5 minutes, and the total number of time slices is 34,272.
    \item \textbf{PEMS-BAY} is a traffic speed dataset collected from California Transportation Agencies (CalTrans) Performance Measurement System (PeMS)~\cite{PEMS-BAY}.
    It contains data of 325 sensors in the Bay Area over a period of 6 months from Jan 1st 2017 to May 31th 2017~\cite{2017DCRNN}.
    The traffic information is recorded at the rate of every 5 minutes, and the total number of time slices is 52,116.
    \item \textbf{PEMS04} is a traffic flow dataset also collected from CalTrans PeMS~\cite{PEMS-BAY}.
    It contains data of 307 sensors in the Bay Area over a period of 2 months from Jan 1st 2018 to Feb 28th 2018~\cite{2019ASTGCN}.
    The traffic information is recorded at the rate of every 5 minutes, and the total number of time slices is 16,992.
    
\end{itemize}
The statistical information is summarized in Table \ref{tab:datasets}.
For a fair comparison, we follow the dataset division in previous works.
For METR-LA and PEMS-BAY, we use about 70\% of data for training, 20\% of data for testing, and the remaining 10\% for validation~\cite{2017DCRNN, GWNet}.
For PEMS04, we use about 60\% of data for training, 20\% of data for testing, and the remaining 20\% for validation~\cite{2021ASTGNN, 2019ASTGCN}.

\begin{table}
\setlength{\abovecaptionskip}{0.cm}
% \setlength{\belowcaptionskip}{-1cm}
% \setstretch{1.18}
\caption{Statistics of datasets.}
\label{tab:datasets}
\scalebox{0.94}{
  \begin{tabular}{c|c|c|c|c}    % Dataset Samples Nodes SampleRate TimeSpan
    \toprule
    \textbf{Dataset} &\textbf{\# Samples} & \textbf{\# Node} & \textbf{Sample Rate} & \textbf{Time Span}\\
    % \midrule
    \midrule
    {METR-LA}  & 34272 & 207 &5mins & 4 months\\
    {PEMS-BAY} & 52116 & 325 &5mins & 6 months\\
    {PEMS04}   & 16992 & 307 &5mins & 2 months\\
    \bottomrule
  \end{tabular}
  }
  \vspace{-0.3cm}
\end{table}

\noindent\textbf{Baselines.} We select a wealth of baselines that have official public code.
% including the traditional methods and the typical deep learning methods, as well as  the very recent state-of-the-art works.
Historical Average~(HA), VAR~\cite{VAR}, and SVR~\cite{SVR} are traditional methods.
FC-LSTM~\cite{2014Seq2Seq}, DCRNN~\cite{2017DCRNN}, Graph WaveNet~\cite{GWNet}, ASTGCN~\cite{2019ASTGCN}, and STSGCN~\cite{2020STSGCN} are typical deep learning methods. GMAN~\cite{2020GMAN}, MTGNN~\cite{2020MTGNN}, and GTS~\cite{2021GTS} are recent state-of-the-art works.
More details of baselines can be found in Appendix \ref{appendix_basline}.

\noindent\textbf{Metrics.}
We evaluate the performances of all baselines by three commonly used metrics in multivariate time series forecasting, including Mean Absolute Error (MAE), Root Mean Squared Error (RMSE) and Mean Absolute Percentage Error (MAPE).

\noindent\textbf{Implementation.}
% The proposed model is implemented by Pytorch 1.9.1 on two NVIDIA Tesla V100 GPUs.
We set patch size $L$ to 12.
We set the number of patches $P$ to 168 for METR-LA and PEMS-BAY, and 336 for PEMS04, \ie we use historical information for a week for METR-LA and PEMS-BAY, and two weeks for PEMS04.
We aim at forecasting the next 12 time steps.
The masking ratio $r$ is set to 75\%.
The hidden dimension of the latent representations of TSFormer $d$ is set to 96.
The TSFormer encoder uses 4 layer of Transformer blocks, and the decoder uses 1 layer.
The number of attention heads in Transformer blocks is set to 4.
The hyper-parameter of Graph WaveNet is set to default in their papers~\cite{GWNet}.
For the $k$NN graph $\mathbf{A}^a$, we set $k$ to 10.
We perform significance tests~(t-test with p-value < 0.05) over all the experimental results.

\begin{table*}[htpb]

\renewcommand\arraystretch{0.98}

    \centering
    \setlength{\abovecaptionskip}{0.cm}
    \caption{Multivariate time series forecasting on the METR-LA, PEMS-BAY, and PEMS04 datasets. Numbers marked with $^*$ indicate that the improvement is statistically significant compared with the best baseline~(t-test with p-value$<0.05$).}
    \label{tab:main}
    \begin{tabular}{ccccr|ccr|ccr}
      \toprule
      \midrule
      \multirow{2}*{\textbf{Datasets}} &\multirow{2}*{\textbf{Methods}} & \multicolumn{3}{c}{\textbf{Horizon 3}} & \multicolumn{3}{c}{\textbf{Horizon 6}}& \multicolumn{3}{c}{\textbf{Horizon 12}}\\ 
      \cmidrule(r){3-5} \cmidrule(r){6-8} \cmidrule(r){9-11}
      &  & MAE & RMSE & MAPE & MAE & RMSE & MAPE & MAE & RMSE & MAPE\\
      \midrule
      \midrule
      \multirow{14}*{\textbf{METR-LA}} 
      &HA              & 4.79  & 10.00 & 11.70\%       & 5.47  & 11.45 & 13.50\%      & 6.99  & 13.89  & 17.54\% \\ 
      &VAR             & 4.42  & 7.80  & 13.00\%       & 5.41  & 9.13  & 12.70\%      & 6.52  & 10.11 & 15.80\% \\ 
      &SVR             & 3.39  & 8.45  & 9.30\%        & 5.05  & 10.87 & 12.10\%      & 6.72  & 13.76 & 16.70\% \\ 
      &FC-LSTM         & 3.44  & 6.30  & 9.60\%        & 3.77  & 7.23  & 10.09\%      & 4.37  & 8.69  & 14.00\% \\ 
      &DCRNN           & 2.77  & 5.38  & 7.30\%        & 3.15  & 6.45  & 8.80\%       & 3.60  & 7.60  & 10.50\% \\ 
      &STGCN           & 2.88  & 5.74  & 7.62\%        & 3.47  & 7.24  & 9.57\%       & 4.59  & 9.40  & 12.70\% \\ 
      &Graph WaveNet   & 2.69  & 5.15  & 6.90\%        & 3.07  & 6.22  & 8.37\%       & 3.53  & 7.37  & 10.01\% \\
      &ASTGCN          & 4.86  & 9.27  & 9.21\%        & 5.43  & 10.61 & 10.13\%      & 6.51  & 12.52 & 11.64\% \\  
      &STSGCN          & 3.31  & 7.62  & 8.06\%        & 4.13  & 9.77  & 10.29\%      & 5.06  & 11.66 & 12.91\% \\  
      &GMAN            & 2.80  & 5.55  & 7.41\%        & 3.12  & 6.49  & 8.73\%       & 3.44  & 7.35  & 10.07\% \\  
      &MTGNN           & 2.69  & 5.18  & 6.88\%        & 3.05  & 6.17  & 8.19\%       & 3.49  & 7.23  & 9.87\% \\  
      &GTS             & 2.67  & 5.27  & 7.21\%        & 3.04  & 6.25  & 8.41\%       & 3.46  & 7.31  & 9.98\% \\  
    \cmidrule(r){2-11}
    &STEP      & \textbf{2.61}$^*$  & \textbf{4.98}$^*$  & \textbf{6.60\%}$^*$        & \textbf{2.96}$^*$  & \textbf{5.97}$^*$  & \textbf{7.96\%}$^*$      & \textbf{3.37}$^*$  & \textbf{6.99}$^*$  & \textbf{9.61\%}$^*$ \\ 
    \midrule
    \midrule
    \multirow{14}*{\textbf{PEMS-BAY}} 
      &HA              & 1.89  & 4.30  & 4.16\%        & 2.50  & 5.82  & 5.62\%       & 3.31  & 7.54  & 7.65\% \\ 
      &VAR             & 1.74  & 3.16  & 3.60\%        & 2.32  & 4.25  & 5.00\%       & 2.93  & 5.44  & 6.50\% \\ 
      &SVR             & 1.85  & 3.59  & 3.80\%        & 2.48  & 5.18  & 5.50\%       & 3.28  & 7.08  & 8.00\% \\ 
      &FC-LSTM         & 2.05  & 4.19  & 4.80\%        & 2.20  & 4.55  & 5.20\%       & 2.37  & 4.96  & 5.70\% \\ 
      &DCRNN           & 1.38  & 2.95  & 2.90\%        & 1.74  & 3.97  & 3.90\%       & 2.07  & 4.74  & 4.90\% \\ 
      &STGCN           & 1.36  & 2.96  & 2.90\%        & 1.81  & 4.27  & 4.17\%       & 2.49  & 5.69  & 5.79\% \\ 
      &Graph WaveNet   & 1.30  & 2.74  & 2.73\%        & 1.63  & 3.70  & 3.67\%       & 1.95  & 4.52  & 4.63\% \\
      &ASTGCN          & 1.52  & 3.13  & 3.22\%        & 2.01  & 4.27  & 4.48\%       & 2.61  & 5.42  & 6.00\% \\  
      &STSGCN          & 1.44  & 3.01  & 3.04\%        & 1.83  & 4.18  & 4.17\%       & 2.26  & 5.21  & 5.40\% \\  
      &GMAN            & 1.34  & 2.91  & 2.86\%        & 1.63  & 3.76  & 3.68\%       & 1.86  & 4.32  & 4.37\% \\  
      &MTGNN           & 1.32  & 2.79  & 2.77\%        & 1.65  & 3.74  & 3.69\%       & 1.94  & 4.49  & 4.53\% \\  
      &GTS             & 1.34  & 2.83  & 2.82\%        & 1.66  & 3.78  & 3.77\%       & 1.95  & 4.43  & 4.58\% \\  
    \cmidrule(r){2-11}
      &STEP      & \textbf{1.26}$^*$  & \textbf{2.73}$^*$  & \textbf{2.59\%}$^*$        & \textbf{1.55}$^*$  & \textbf{3.58}$^*$  & \textbf{3.43\%}$^*$      & \textbf{1.79}$^*$  & \textbf{4.20}$^*$  & \textbf{4.18\%}$^*$ \\ 
    \midrule  
    \midrule
    \color{black}{\multirow{14}*{\textbf{PEMS04}}}
      &HA              & 28.92  & 42.69  & 20.31\%        & 33.73  & 49.37  & 24.01\%       & 46.97  & 67.43  & 35.11\% \\ 
      &VAR             & 21.94  & 34.30  & 16.42\%        & 23.72  & 36.58  & 18.02\%        & 26.76  & 40.28  & 20.94\% \\ 
      &SVR             & 22.52  & 35.30  & 14.71\%        & 27.63  & 42.23  & 18.29\%       & 37.86  & 56.01  & 26.72\% \\ 
      &FC-LSTM         & 21.42  & 33.37  & 15.32\%        & 25.83  & 39.10  & 20.35\%       & 36.41  & 50.73  & 29.92\% \\ 
      &DCRNN           & 20.34  & 31.94  & 13.65\%        & 23.21  & 36.15  & 15.70\%       & 29.24  & 44.81  & 20.09\% \\ 
      &STGCN           & 19.35  & 30.76  & 12.81\%        & 21.85  & 34.43  & 14.13\%       & 26.97  & 41.11  & 16.84\% \\ 
      &Graph WaveNet   & 18.15  & 29.24  & 12.27\%        & 19.12  & 30.62  & 13.28\%       & 20.69  & 33.02  & 14.11\% \\
      &ASTGCN          & 20.15  & 31.43  & 14.03\%        & 22.09  & 34.34  & 15.47\%       & 26.03  & 40.02  & 19.17\% \\  
      &STSGCN          & 19.41  & 30.69  & 12.82\%        & 21.83  & 34.33  & 14.54\%       & 26.27  & 40.11  & 14.71\% \\  
      &GMAN            & 18.28  & 29.32  & 12.35\%        & 18.75  & 30.77  & 12.96\%       & 19.95  & \textbf{30.21}  & 12.97\% \\  
      &MTGNN           & 18.22  & 30.13  & 12.47\%        & 19.27  & 32.21  & 13.09\%       & 20.93  & 34.49  & 14.02\% \\  
      &GTS            & 18.97  & 29.83  & 13.06\%        & 19.29  & 30.85  & 13.92\%       & 21.04  & 34.81  & 14.94\% \\  
    \cmidrule(r){2-11}
      &STEP      & \textbf{17.34}$^*$  & \textbf{28.44}$^*$  & \textbf{11.57\%}$^*$        & \textbf{18.12}$^*$  & \textbf{29.81}$^*$  & \textbf{12.00\%}$^*$      & \textbf{19.27}$^*$  & 31.33  & \textbf{12.78\%}$^*$ \\ 
        \midrule
      \bottomrule
    \end{tabular}
  \end{table*}

% \vspace{-0.3cm}
\subsection{Main Results}
As shown in Table \ref{tab:main}, our STEP framework consistently achieves the best performance in almost all horizons in all datasets, indicating the effectiveness of our framework.
GTS and MTGNN jointly learn the graph structure among multiple time series and the spatial-temporal graph neural networks.
GTS extends DCRNN by introducing a neighborhood graph as a regularization to improve graph quality and reformulates the problem as a unilevel optimization problem. 
MTGNN replaces the GNN and Gated TCN in Graph WaveNet with mix-hop propagation layer~\cite{2019MixHop} and dilated inception layer, and proposes to learn latent adjacency matrix to seek further improvement.
However, they can not consistently outperform other baselines.
Kindly note that the results of GTS may have some gaps with the original paper because it calculates the evaluation metrics in a slightly different manner.
Some details can be found in the appendix in the original paper~\cite{2021GTS} and similar issues in its official code repository\footnote{\url{https://github.com/chaoshangcs/GTS/issues}}.
We unify the evaluation process with other baselines, run GTS five times, and report its best performance.
GMAN performs better in long-term prediction benefiting from the powerful ability of the attention mechanism in capturing long-term dependency.
DCRNN and Graph WaveNet are two typical spatial-temporal graph neural networks. 
Even compared with many newer works such as ASTGCN and STSGCN, their performance is still very promising. 
This may be due to their refined and reasonable model architecture. 
FC-LSTM, a classic recurrent neural network, can not perform well since it only considers temporal features, ignoring the dependencies between time series.
Other non-deep learning methods HA, VAR, and SVR perform worst since they have strong assumptions about the data, \eg stationary or linear. 
Thus, they can not capture the strong nonlinear and dynamic spatial and temporal correlations in real-world datasets.

In a nutshell, STEP provides stable performance gains for Graph WaveNet by fully exploiting representations extracted by TSFormer from very long-term historical time series.
However, despite the significant performance improvements, it is difficult for us to intuitively understand what TSFormer has learned and how it can help STGNNs. 
% In the next subsection, we will verify that TSFormer has learned multiple periodicities in the time dimension.
In the next subsection, we will inspect the TSFormer and demonstrate the learned multiple periodicities temporal pattern.

\subsection{Inspecting The TSFormer}
\label{sec_inspecting}

\begin{figure}[t]
    \setlength{\abovecaptionskip}{0.2cm}
    \setlength{\belowcaptionskip}{-0.4cm}
  \centering
  \includegraphics[width=0.97\linewidth]{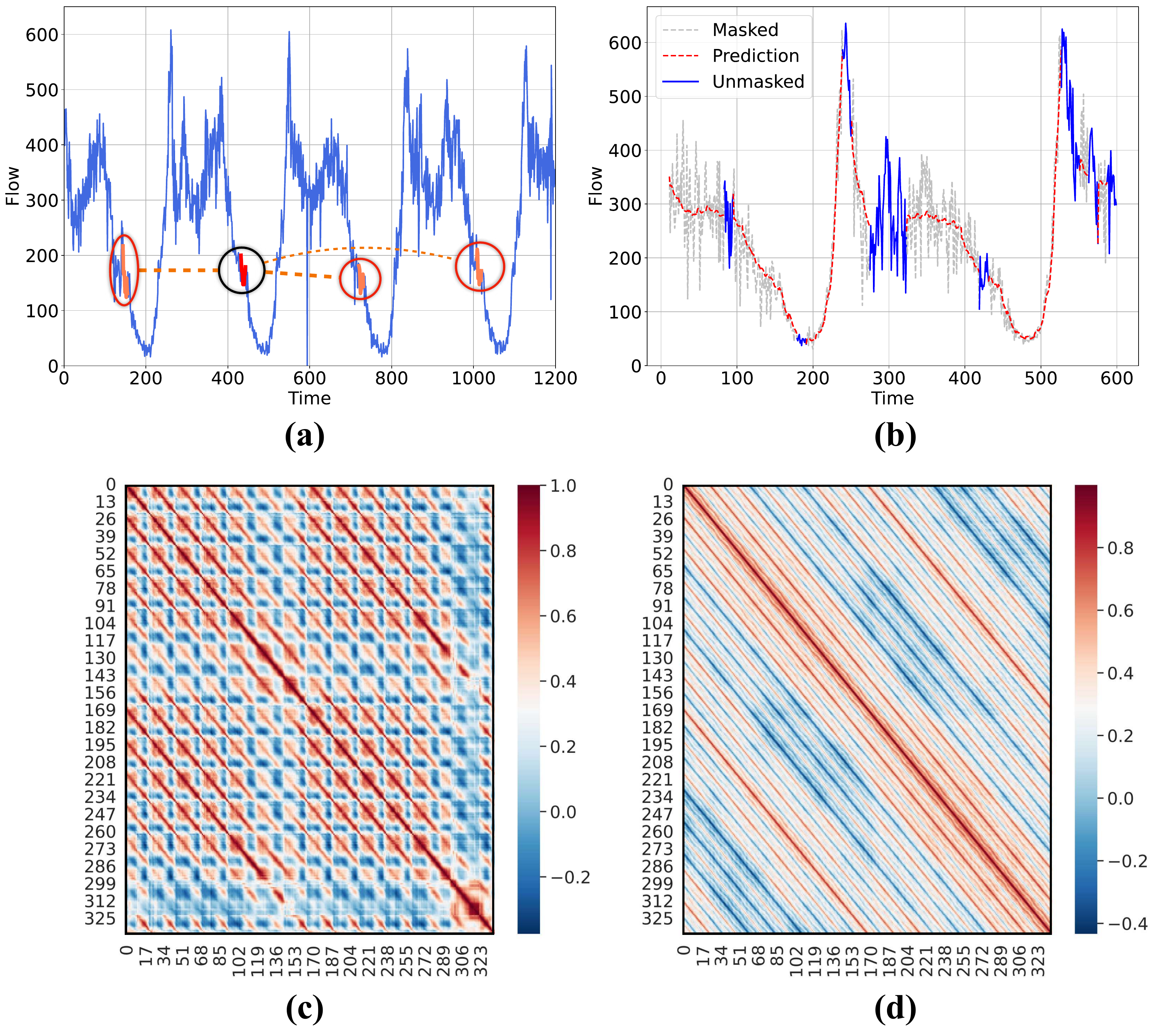}
  \caption{Inspecting the TSFormer. (a) Learned temporal periodicity. (b) Reconstruction. (c) Similarity of  latent representations among different patches. (d) Similarity of positional embeddings among different patches.}
  \label{inspecting}
\end{figure}

% 目标：在这一部分，我们想要直观的理解TSFormer学到了什么。
% 设置：因此，我们基于PEMS04数据集，随机选取一个时间序列并随机选取一个验证样本，来分析TSFormer及其输出。在我们的设置中，PEMS04一个样本具有336个Patches，即覆盖两个周的时间长度。
In this subsection, we would like to intuitively explore what TSFormer has learned.
We conduct experiments on the PEMS04 dataset.
Specifically, we randomly select a time series in the PEMS04 and then randomly choose a sample of the test dataset to analyze TSFormer.
Note that each input sample in PEMS04 has 336 patches of length 12, which means it covers data of the past two weeks.

% \subsubsection{Learned Temporal Pattern}
\noindent\textbf{Learned temporal pattern.}
% 第一个我们想要知道的问题是TSFormer是否学到时间维度的模式。
% 我们期望它输出有意义的表征，并且能够解决图1(b)中的问题。
% 因此，给定输入的Patch，我们随机选取其中一个，然后计算输出的表征和其他Patches的相似度，选取最相似的几个。
% 我们使用Cos Similarity作为相似度度量。
% 实验结果如图()所示。所有selected paches都用红色的线画出。我们选择了第36个Patches作为匹配模板，我们选取3个相似的其他Patches，which被红色圈出。显然模型正确的归类Patches。
% 为了获取bigger picture，我们还计算了所有Patches两两之间的相似度，得到了一个
% 其结果如图()所示，呈现了明显的周期性。
% Patches之间的相似性由daily模式和Weekly模式决定。
% 每一个Patch会一天中同一时刻的Patch相似，和一周中同一天的同一时刻的Patches更相似。
% 全蓝的Column或者Row，代表这个时刻传感器出现了宕机或者存在很大的噪声波动，造成它和其他的Patches都不相似。
% 由于TSFormer学到了正确的Patches之间的关系，它能够显著地增强下游时空图神经网络也就不奇怪了。
Firstly, we would like to explore whether TSFormer learned temporal patterns.
We expect it to generate meaningful representations and be able to solve the problem in Figure \ref{Intro}(b).
Therefore, we randomly select a patch, compute the cosine similarity with the representations of all the other patches, and select the most similar 3 patches. The result is shown in Figure \ref{inspecting}(a), where the original patch is in the black circle, and the selected most similar patches are in the red circle. 
% Apparently, TSFormer extract has learned temporal patterns.
Apparently, TSFormer has a strong ability to identify similar patches.
Furthermore, in order to get the bigger picture, we also calculate the pairwise similarity between all patches and get a $336\times 336$ heat map, where element in $i$-th column and $j$-th row indicates the cosine similarity between patch $i$ and patch $j$.
The result shown in Figure \ref{inspecting}(c) presents clear daily and weekly periodicities.
% Each patch is similar to a patch at the same time of a day, and more similar to a patch at the same time of the week on the same day.
For each patch, it is similar to the patch at the same time of a day, and the most similar patch usually falls on the same time of the week on the same day. 
The observation is in line with human intuition.
The blue columns or rows mean that the sensor is down or has a large noise fluctuation at this moment, which makes it different from other patches.
Since TSFormer has learned the correct relationship between patches, it is reasonable that it can significantly enhance the downstream STGNNs.

% \subsubsection{Reconstruction Visualization}
\noindent\textbf{Reconstruction visualization.}
% 另外，我们还可视化了TSFormer重建的结果。
% 结果显示TSFormer能够基于少量的Unmasked Patches准确的重建出Masked Patches。
Additionally, we also visualized the results of the TSFormer reconstruction, which is shown in Figure \ref{inspecting}(b), where the grey line presents masked patches and the red line demonstrates the reconstruction.
The results show that TSFormer can effectively reconstruct masked patches based on a small number of unmasked patches~(blue line).

% \subsubsection{Positional embeddings}
\noindent\textbf{Positional embeddings.}
% TSFormer和MAE、经典的Transformer之间的一个较大的不同就是可学习的Positional Embedding。
% 我们想要知道它是否学习到了合理的位置编码。
% 因此，我们计算336个Patches的位置编码两两之间的Cosine相似度，其热力图如图(a)所示。
% 我们惊讶的发现，TSFormer的位置表征非常好的反映了时间序列中的多周期性。
% 主要是因为不同于输出的表征需要依赖于输入，位置编码是完全自由训练的，它更少地受输入数据的噪声的影响。
% TSFormer能够学习到的符合时间模式的位置表征是TSFormer能够成功的关键因素。
% 将可学习的PE替换成sin PE时，我们发现TSFormer无法得到有意义的输出。
Another important difference between TSFormer and MAE~\cite{2021MAE} and the original Transformer~\cite{2017Transformer} is the learnable positional embedding. 
Therefore, we would like to explore whether TSFormer has learned reasonable positional embeddings.
We compute the cosine similarity between the positional embeddings of 336 patches and get a $336\times 336$ heat map, shown in Figure \ref{inspecting}(d).
We find that the positional embedding of TSFormer better reflects the multi-periodicity in time series.
This is because, unlike the representation of the encoder, which needs to depend on the input patches, the positional embeddings are completely free to optimize and are less affected by the noise of the input data.
We conjecture that such positional embedding is the key factor for the success of TSFormer since we found that TSFormer could not get meaningful representations if we replace the learnable positional embeddings with the deterministic, sinusoidal ones.

\vspace{-0.2cm}
\subsection{Ablation Study}
\begin{figure}
    \setlength{\abovecaptionskip}{0.0cm}
    \setlength{\belowcaptionskip}{-0.4cm}
  \centering
  \includegraphics[width=0.95\linewidth]{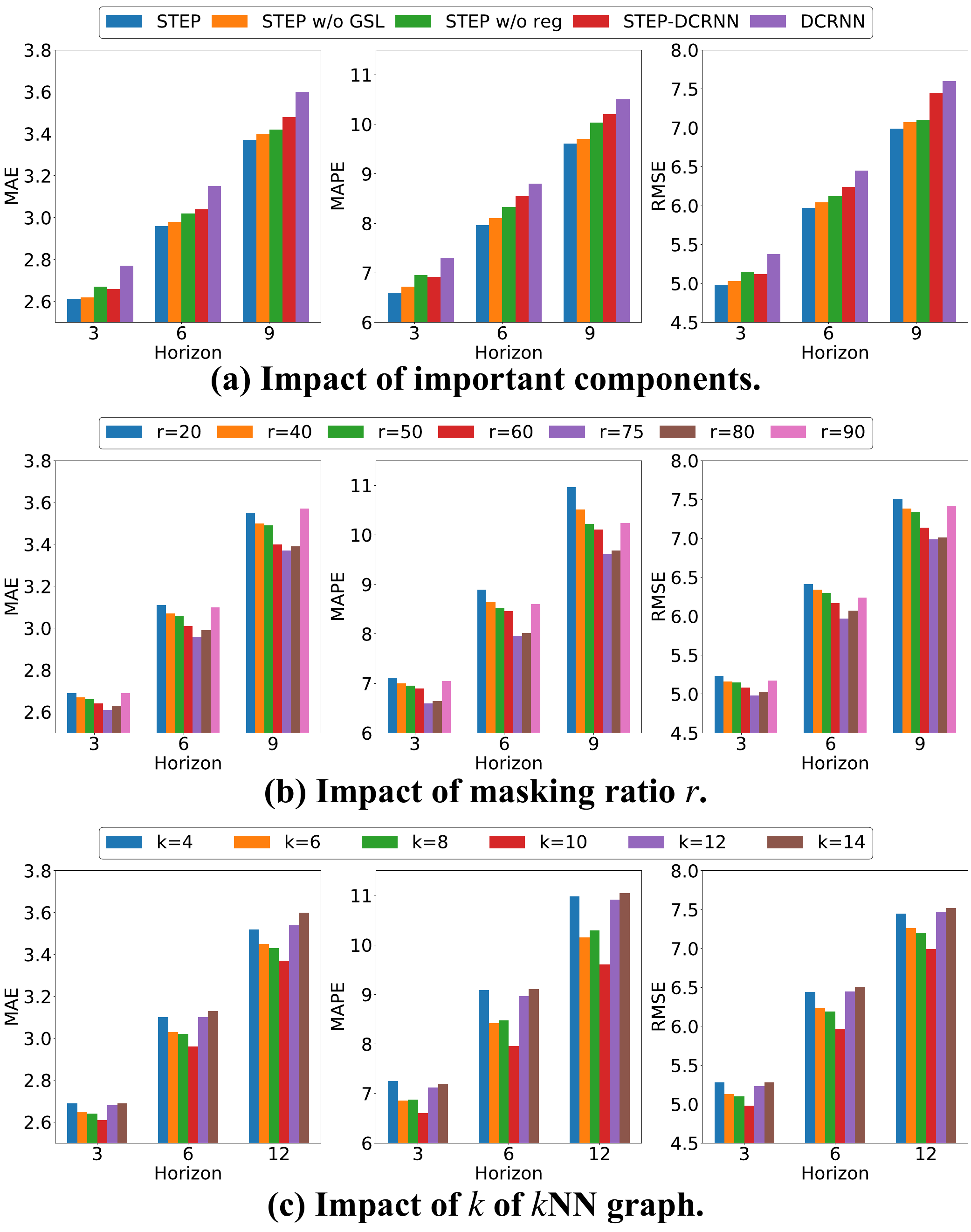}
  \caption{Ablation study and hyper-parameter study.}
  \label{ablation}
\end{figure}

In this part, we conduct experiment to verify the impact of some key components.
First, we set \textit{STEP w/o GSL} to test the performance without the graph structure learning model.
Second, we set \textit{STEP w/o reg} to replace the $k$NN graph computed by the representations of TSFormer with the $k$NN graph in GTS~\cite{2021GTS}, which is computed based on the cosine similarity of raw time series $\mathbf{S}^{i}_{train}$, to test the superiority of the long sequence representations of TSFormer.
Finally, we also test more downstream STGNNs to verify the generality of STEP.
We choose DCRNN as another backend, \ie \textit{STEP-DCRNN}.
Additionally, we also present the performance of DCRNN for comparison.
The results are shown in Figure \ref{ablation}(a).

% STEP w/o GSL若于STEP，我们的GSL模型始终扮演一个正向角色。
% 但STEP w/o的性能依旧很好，这意味着TSFormer产生的segment-level的表征非常非常重要。
% STEP和STEP w/o reg比，意味着TSFormer的表达能力能进一步提升学习到的图的质量。
% 正如Intro中提到的，DCRNN和GWNet是两类最具代表性的方法。DCRNN代表了一大类基于Seq2Seq架构的STGNN[123]。
% 我们按照公式(0)，将TSFormer的表征加到Seq2Seq架构Encoder端的lantent 表征上。
% 可以看到STEP模型显著地增强了DCRNN的性能，这验证了我们的框架的通用性。
As can be seen from the figure, STEP outperforms \textit{STEP w/o GSL}, which shows that our graph structure learning module consistently plays a positive role.
Meanwhile, \textit{STEP w/o GSL} still achieves satisfactory performance, demonstrating that segment-level representation plays a vital role.
STEP also outperforms \textit{STEP w/o reg}, showing that the long sequence representations of TSFormer is superior in improving the graph quality.
In addition, as mentioned in Section \ref{sec_intro}, DCRNN represents a large class of STGNNs~\cite{2019STMetaNet, 2021GTS, 2021REST, 2020GMAN} that are based on the seq2seq~\cite{2014Seq2Seq} architecture.
We fuse the representation of TSFormer to the latent representations of the seq2seq encoder according to Eq.(\ref{fuse}).
We can see that STEP significantly enhances the performance of DCRNN, which verifies the generality of STEP.

\subsection{Hyper-parameter Study}
{\color{black}
We conduct experiments to analyze the impacts of two hyper-parameters: the masking ratio $r$, the $k$ of the $k$NN graph in graph structure learning. We present the results on METR-LA dataset.

The effect of $r$ and $k$ are shown in Figure \ref{ablation}(b) and Figure \ref{ablation}(c), respectively. We find there exist optimal values for both $r$ and $k$. 
For masking ratio $r$, when $r$ is small, masked values in time series can be predicted by simple average or interpolation. Thus it creates a trivial self-supervised learning task and can not get useful representations. 
When $r$ is large, the model would lose too much information and fail to learn temporal patterns. % In the experiment, we find the value of 75\% results in the optimal model performance.
For $k$ of the $k$NN graph in graph structure learning, a small value of $k$ would make the learned graph incomplete and lose dependency information, thus the performance is worse. A large value of $k$ would introduce redundancies, which may hurt the information aggregation of graph neural networks, leading to unsatisfactory performance. }
\section{Related Work}
\subsection{Spatial-Temporal Graph Neural Networks}
The accuracy of multivariate time series forecasting has been largely improved by artificial intelligence~\cite{Innovation}, especially deep learning techniques.
Among these techniques, Spatial-Temporal Graph Neural Networks~(STGNNs) are the most promising methods, 
which combine Graph Neural Networks (GNNs)~\cite{2017GCN, 2016ChebNet} and sequential models~\cite{2014GRU, 2014Seq2Seq} to model the spatial and temporal dependency jointly.
Graph WaveNet~\cite{GWNet}, MTGNN~\cite{2020MTGNN}, STGCN~\cite{2018STGCN}, and StemGNN~\cite{2020StemGNN} combine graph convolutional networks and gated temporal convolutional networks with their variants. These methods are based on convolution operation, which facilitates parallel computation.
DCRNN~\cite{2017DCRNN}, ST-MetaNet~\cite{2019STMetaNet}, AGCRN~\cite{2020AdaptiveGCRN}, and TGCN~\cite{2019TGCN} combine diffusion convolutional networks and recurrent neural networks~\cite{2014GRU, 2014Seq2Seq} with their variants. 
They follow the seq2seq~\cite{2014Seq2Seq} architecture to predict step by step.
Moreover, attention mechanism is widely used in many methods, such as GMAN~\cite{2020GMAN} and ASTGCN~\cite{2019ASTGCN}. 
Although STGNNs have made significant progress, the complexity of STGNNs is high because it needs to deal with both temporal and spatial dependency at every step.
Therefore, STGNNs can only take short-term historical time series as input, such as the past 1 hour~(twelve time steps in many datasets).

More recently, an increasing number of works~\cite{2018NRI, 2019LDS, 2021GTS} have focused on joint learning of graph structures and graph neural networks to model the dependencies between nodes.
% NRI~\cite{2018NRI} takes the form of a variational auto-encoder, in which the latent code represents the underlying dependency graph and the reconstruction is based on graph neural networks.
LDS~\cite{2019LDS} models the edges as random variables whose parameters are treated as hyperparameters in a bilevel learning framework.
The random variables parameterize the element-wise Bernoulli distribution from which the adjacency matrix $\mathbf{A}$ is sampled.
GTS~\cite{2021GTS} introduces a neighborhood graph as a regularization that improves graph quality and reformulates the problem as a unilevel optimization problem.
Notably, We follow the framework of GTS but enhance it by the pre-training model since TSFormer gives better latent representations of time series for calculating their correlations.
% We will conduct the ablation studies in Section {\color{red}X}.

% GWNet MTGNN STGCN 2020StemGNN
% DCRNN STMetaNet 2020AdaptiveGCRN 2019TGCN
% GMAN 
% FCGAGA 2021HGCN

% Spatial Temporal Graph Neural Networks: DCRNN, GWNet, MTGNN, ASTGCN, STGCN, GMAN, 2020StemGNN. FCGAGA, 2020AdaptiveGCRN, 2021HGCN, 2019STMetaNet, 2019TGCN, Innovation, GCN
% Graph Stucture Learning: NRI, LDS, GTS, 2021REST
% 结论：取得了很大的进展，但是由于复杂度的限制，只能应用在短期序列上；
\subsection{Pre-training Model}
% NLP
The pre-training model is used to learn a good representation from massive unlabeled data and then use these representations for other downstream tasks.
Recent studies have demonstrated significant performance gains on many natural language processing tasks with the help of the representation extracted from pre-training models~\cite{2020PTM}.
Prominent examples are the BERT~\cite{2019BERT} and GPT~\cite{2020GPT}, which are based on the Transformer encoder and decoder, respectively.
The Transformer architecture is more powerful and more efficient than LSTM architecture~\cite{2014Seq2Seq, 2018ELMO} and has become the mainstream approach for designing pre-training models.
% CV
More recently, Transformer for images has attracted increasing attention because of its powerful performance.
ViT~\cite{2020ViT} proposes to split an image into patches and provide the sequence of linear embeddings of these patches as an input to a Transformer, showing impressive performance.
However, ViT needs supervised training, which requires massive labeled data.
On the contrary, MAE~\cite{2021MAE} uses self-supervised learning based on the masked autoencoding strategy. 
MAE enables us to train large models efficiently and effectively and outperforms supervised pre-training.
Although the pre-training model has made significant progress in natural language processing and computer vision, progress in time series lags behind them. In this paper, we propose a pre-training model~(named TSFormer) for time series based on Transformer blocks and improve the performance of the downstream forecasting task.

\section{Conclusion}
In this paper, we propose a novel STEP framework for multivariate time series forecasting to address the inability of STGNNs to learn long-term information.
The downstream STGNN is enhanced by a scalable time series pre-training model TSFormer.
TSFormer is capable of efficiently learning the temporal pattern from very long-term historical time series and generating segment-level representations, which provide rich contextual information for short-term input of STGNNs and facilitate modeling dependencies between time series.
Extensive experiments on three real-world datasets show the superiority of the STEP framework and the proposed TSFormer.

\begin{acks}
This work is partially supported by NSFC No. 61902376 and No. 61902382.
In addition, Zhao Zhang is supported by the China Postdoctoral Science Foundation under Grant No. 2021M703273.
\end{acks}

\bibliographystyle{ACM-Reference-Format}
\normalem
\bibliography{references_dblp}
\appendix
\section{More Experiments Details}
\label{appendix_experiments}
\subsection{Baseline Details}
\label{appendix_basline}
\begin{itemize}
    \item \textbf{HA}: 
    Historical Average model, which models time series as a periodic process and uses weighted averages from previous periods as predictions for future periods.
    \item \textbf{VAR}: 
    Vector Auto-Regression~\cite{VAR, lutkepohl2005new} assumes that the past time series is stationary and estimates the relationship between the time series and their lag value.~\cite{2020STGNN}
    \item \textbf{SVR}: Support Vector Regression (SVR) uses linear support vector machine for classical time series regression task.
    \item \textbf{FC-LSTM}~\cite{2014Seq2Seq}: Long Short-Term Memory network with fully connected hidden units is a well-known network architecture that is powerful in capturing sequential dependency.
    \item \textbf{DCRNN}~\cite{2017DCRNN}: Diffusion Convolutional Recurrent Neural Network~\cite{2017DCRNN} replaces the fully connected layer in GRU~\cite{2014GRU} by diffusion convolutional layer to form a new Diffusion Convolutional Gated Recurrent Unit~(DCGRU).
    \item \textbf{Graph WaveNet}~\cite{GWNet}: Graph WaveNet stacks gated temporal convolutional layer and GCN layer by layer to jointly capture the spatial and temporal dependencies.
    \item \textbf{ASTGCN}~\cite{2019ASTGCN}: ASTGCN combines the spatial-temporal attention mechanism to capture the dynamic spatial-temporal characteristics simultaneously.
    \item \textbf{STSGCN}~\cite{2020STSGCN}: STSGCN is proposed to effectively capture the localized spatial-temporal correlations and consider the heterogeneity in spatial-temporal data.
    \item \textbf{MTGNN}~\cite{2020MTGNN}: MTGNN extends Graph WaveNet through the mix-hop propagation layer in the spatial module, the dilated inception layer in the temporal module, and a more delicate graph learning layer.
    \item \textbf{GMAN}~\cite{2020GMAN}: GMAN is an attention-based model which stacks spatial, temporal and transform attentions.
    \item \textbf{GTS}~\cite{2021GTS}: GTS learns a graph structure among multiple time series and forecasts them simultaneously with DCRNN. 
\end{itemize}

\subsection{Optimization Settings}

\begin{table}[h]
\caption{Pre-training setting.}
\label{tab_pretrain}
\centering  
\begin{tabular}{p{3cm}|p{4cm}}
config  & value \\
\toprule
optimizer & AdamW~\cite{AdamW}\\
base learning rate & 5.0e-4\\
weight decay & 0\\
epsilon & 1.0e-8\\
optimizer momentum & $\beta_1, \beta_2=0.9, 0.95$\\
learning rate schedule & MultiStepLR\\
milestones & 50\\
gamma & 0.5\\
gradient clip & 5\\ 
\end{tabular}
\end{table}

\noindent \textbf{Pre-training stage.} 
The default setting is shown in Table \ref{tab_pretrain}. 
We use uniform distribution to initialize the positional embeddings, and we use truncated normal distribution with $\mu=0$ and $\sigma=0.02$ to initialize the mask token, similar to MAE~\cite{2021MAE}. 
We use PyTorch official implementation to implement the Transformer blocks.
We use the linear scaling rule for learning rate and batch size: lr = base\_lr $\times$ (batch\_size/8) for all datasets in the pre-training stage.

\begin{table}[h]
\caption{Forecasting setting.}
\label{tab_forecast}
\centering  
\begin{tabular}{p{3cm}|p{4cm}}
config  & value \\
\toprule
optimizer & Adam~\cite{Adam}\\
learning rate & 0.001/0.005/0.002\\
&{\small(PEMS-BAY/METR-LA/PEMS04)}\\
batch size & 64/64/32\\
&{\small(PEMS-BAY/METR-LA/PEMS04)}\\
weight decay & 1.0e-5\\
epsilon & 1.0e-8\\
learning rate schedule & MultiStepLR\\
milestones & [1, 18, 36, 54, 72]\\
gamma & 0.5\\
gradient clip & 5\\
\text{cl\_num}&3\\
\text{warm\_num}&30\\
\end{tabular}
\end{table}

\noindent \textbf{Forecasting stage.} 
All the settings are shown in Table \ref{tab_forecast}.
Following many recent works, such as MTGNN~\cite{2020MTGNN} and GTS~\cite{2021GTS}, we use the curriculum learning strategy for the forecasting task.
The strategy gradually increases the prediction length of the model with the increase in iteration number.
We increase the prediction length by one per \text{cl\_num} epochs.
Moreover, we additionally perform a warm-up of \text{warm\_num} epochs to better initialize the model for curriculum learning.
In addition, $\lambda$ in Equation (\ref{full_loss}) decays by $\lambda=1/(\lceil epoch/6 \rceil)$, where $\lceil \cdot \rceil$ means ceiling function and \textit{epoch} is the epoch number.

% Mask
% 
% 输入输出
% 维度参数
% 优化器参数
% 优化步骤参数
% Number of Head 

\begin{figure}[ht]
    \setlength{\abovecaptionskip}{0.2cm}
  \centering
  \includegraphics[width=1\linewidth]{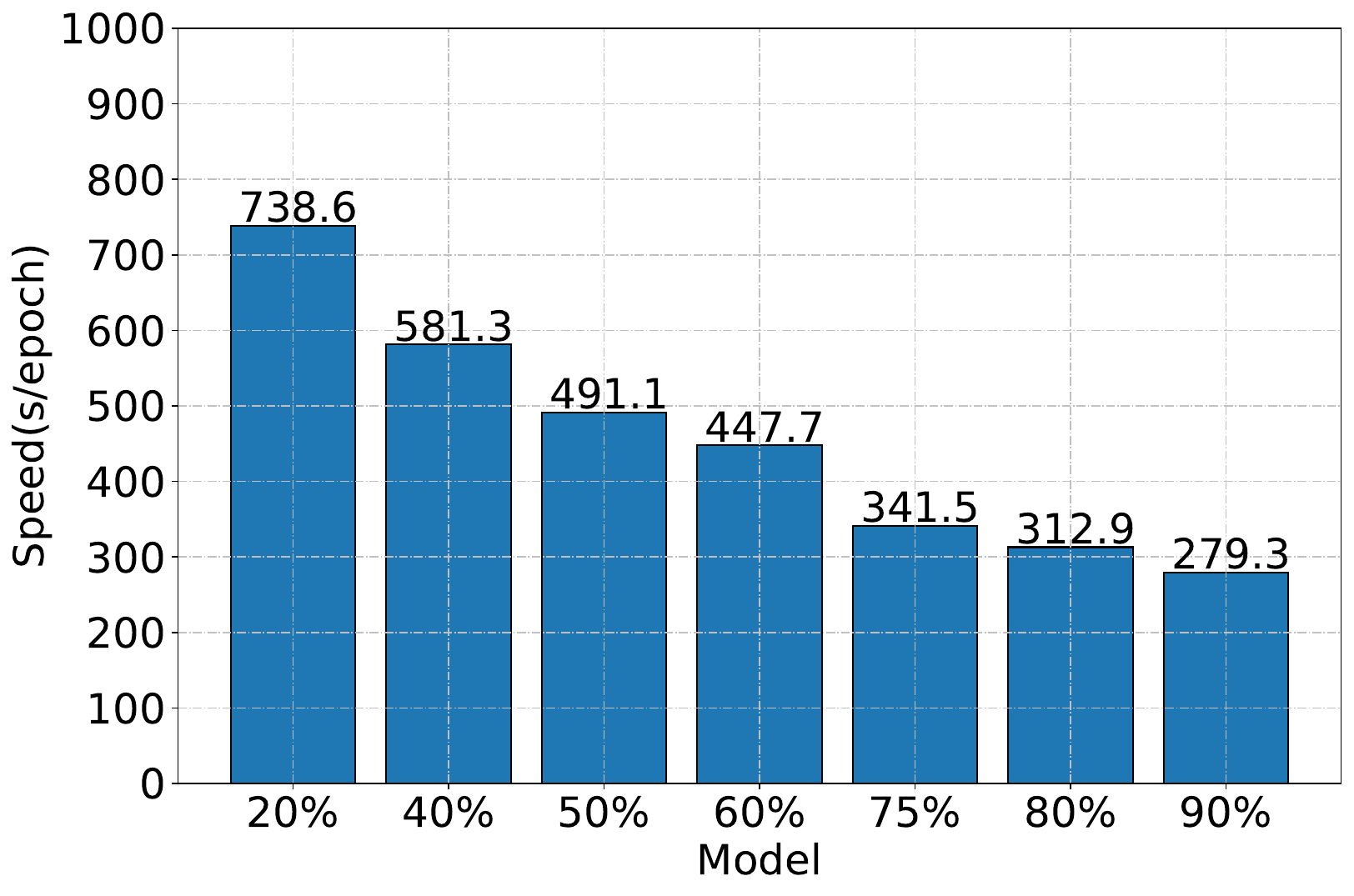}
  \caption{Training speed of different masking ratio $r$.}
  \label{speed}
\end{figure}

\begin{figure}[ht]
    \setlength{\abovecaptionskip}{0.2cm}
  \centering
  \includegraphics[width=1\linewidth]{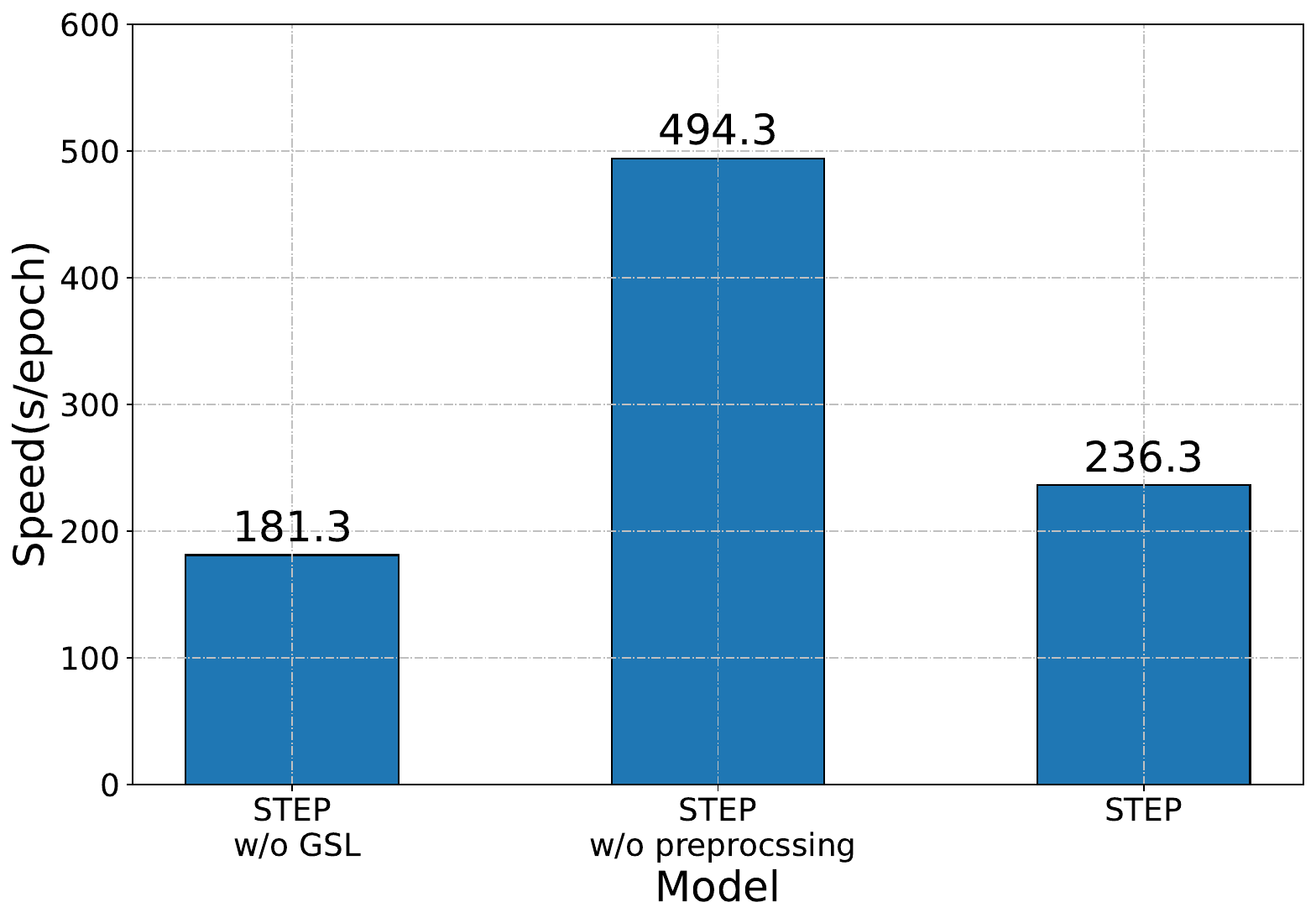}
  \caption{Training speed of different methods.}
  \label{speed2}
\end{figure}

\section{Efficiency}
\label{appendix_efficiency}
% 分为预训练和预测两部分
% 预训练我们对比不同Mask Ratio下的训练时间
% 测试阶段我们对比几个Baseline的训练时间。
% % 测试阶段模型有两个结果，一个是with preprocess，另一个是w/o preprocess
In this part, we compare the efficiency of STEP with other models and their own variants based on the METR-LA dataset.
For a more intuitive and effective comparison, we compare the average training time required for each epoch.
All the experiments are running on an Intel(R) Xeon(R) Gold 5217 CPU @ 3.00GHz, 128G RAM computing server, equipped with RTX 3090 graphics cards.
First, we compare the efficiency of TSFormer in the pre-training stage under different masking ratios.
The result is shown in Figure \ref{speed}.
As the masking ratio increases, the TSFormer will be more efficient.
In summary, thanks to the high masking ratio and fewer Transformer blocks in the encoder and decoder, the TSFormer is lightweight and can be trained efficiently on a single NVIDIA 3090 GPU.

Second, we compare the efficiency of STEP framework in the forecasting stage with its variants.
Recalling that the parameter of TSFormer is fixed during the forecasting stage, we can use TSFormer to provide off-the-shelf representations by preprocessing the whole dataset to reduce redundant calculations in the training process.
We also test the efficiency of STEP without preprocessing, denoting the variant as \textit{STEP w/o pre}.
In addition, we test the efficiency of STEP without graph structure learning, \ie \textit{STEP w/o GSL}.
% Recalling that the pre-trained TSFormer is not trainable in the forecasting stage, we can preprocess all the samples and generate the latent representations in advance. 
% We also test the efficiency of STEP without preprocessing, denoting the variant as \textit{STEP w/o pre}.
The result is shown in Figure \ref{speed2}.
% 预训练确实大大地减少了重复计算
% 但我们发现IO的消耗可能是另一个瓶颈：因为我们需要频繁地从磁盘上读取包含着XXX和XXX的文件并将他们送入到GPU。在我们的实验中，数据读取时间大概占到1/3左右，这个数值随着硬盘缓存的增大会有所改善。
We have the following findings: 
(i) the graph structure learning module accounts for about 55s per epoch on average.
(ii) preprocessing does significantly reduce repetitive computations.
% But we found that disk I/O may be another bottleneck because we need to frequently read files containing $\mathbf{H}^i$ from disk and send them to GPU. In our experiment, the data fetching time accounts for about 1/3 of the total time, and this value will decrease with the increase of the SSD cache and speed.

% While STEP doesn't add much complexity after preprocessing, disk I/O can become the bottleneck because we need to read large files storing $\mathbf{H}^i$from disk and send them to GPU frequently. 
% The yellow part shows the time to fetch data, which is nearly half of the total time. 
% Therefore, how to further avoid or alleviate the disk I/O bottleneck is an important issue that need to be addressed in the further.

% Although STEP does not add much more complexity after preprocessing, the disk IO may become the bottleneck since we need to frequently read large files that store $\mathbf{H}^i$ from disk.
% The yellow part demonstrates the data fetching time, which accounts for nearly half of the total time.
% Therefore, how to further break the disk bottleneck is a important issue.

\section{Visualization}
\label{appendix_visualization}
\begin{figure}[ht]
  \centering
  \includegraphics[width=1\linewidth]{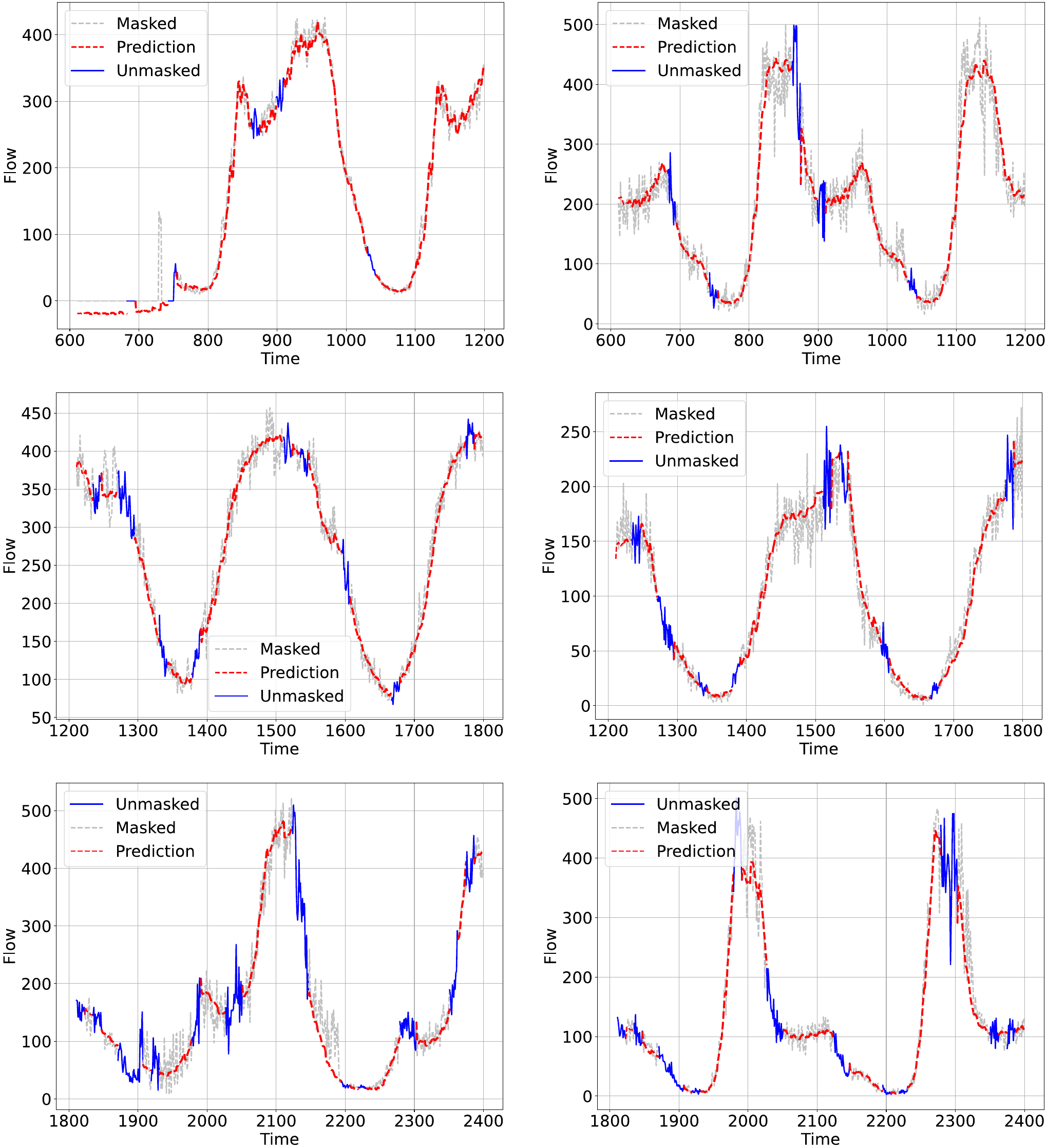}
  \caption{Reconstruction visualizations.}
  \label{vis_rec}
\end{figure}

\begin{figure}[ht]
  \centering
  \includegraphics[width=1\linewidth]{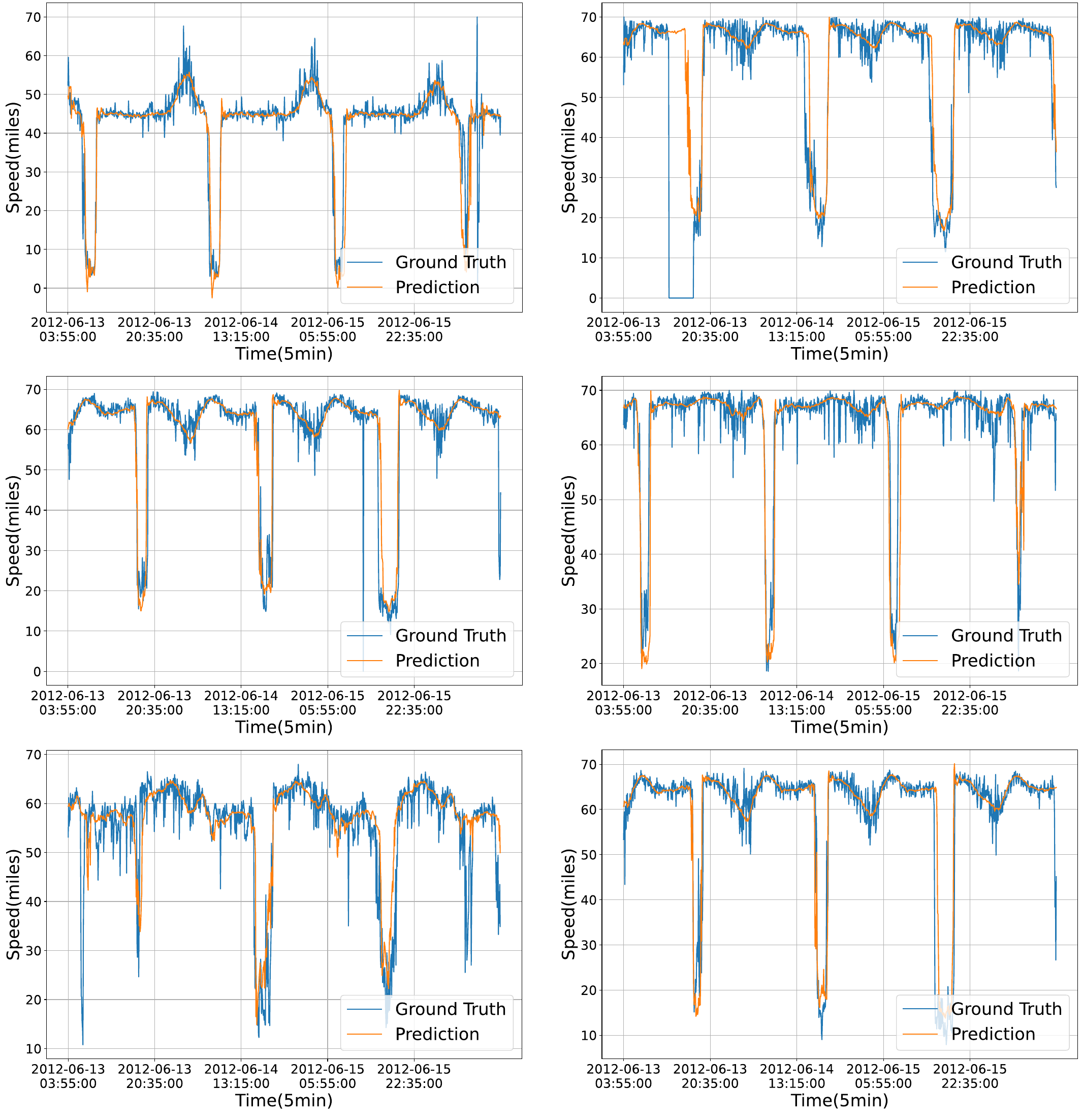}
  \caption{Forecasting visualizations.}
  \label{vis_for}
\end{figure}

In order to further intuitively understand and evaluate our model, in this section, we give more visualizations.
First, we provide more visualizations about reconstructions of the TSFormer on PEMS04 dataset like Figure \ref{inspecting}(b).
The results are shown in Figure \ref{vis_rec}. 
Note that due to space limitation, we only visualize time series in a small window rather than the whole input time series $\mathbf{S}^i$.
% We are surprisingly find that even given very limited information surrounding the unmasked patches, TSFormer reconstructs the masked patches very accurately.
Surprisingly, we find that even given very limited information surrounding the unmasked patches, TSFormer reconstructs the masked patches accurately.
These results again indicate that our model has a strong ability to learn rich temporal patterns from very long-term time series.
Then, we visualize the prediction of our model and the groundtruth data based on METR-LA dataset.
We randomly selected six time series and displayed their data from June 13th 2012 to June 16th 2012~(located the test dataset).
The forecasting results on six randomly selected time series are shown in Figure \ref{vis_for}.
We can see that our model can accurately make predictions for different time series.
Furthermore, we find that the model has the ability to resist noise. 
For example, in the right top figure, the traffic sensor apparently failed in the afternoon of June 13th, 2012. 
However, the model does not overfit the noise.

% gives more visualization about the reconstruction shown in \ref{inspecting}(b) and prediction of our STEP.

% visualize the prediction of our STEP and the real data. 
% Furthermore, we demonstrate more reconstruction results of TSFormer.
% In order to further intuitively understand and evaluate our model, in this section, we visualize the prediction of our STEP and the real data. 
% Furthermore, we demonstrate more reconstruction results of TSFormer.
% In order to further intuitively understand and evaluate our model, in this section, we visualize the prediction of our STEP and the real data. 
% Furthermore, we demonstrate more reconstruction results of TSFormer.
% In order to further intuitively understand and evaluate our model, in this section, we visualize the prediction of our STEP and the real data. 
% Furthermore, we demonstrate more reconstruction results of TSFormer.
% In order to further intuitively understand and evaluate our model, in this section, we visualize the prediction of our STEP and the real data. 
% Furthermore, we demonstrate more reconstruction results of TSFormer.
% In order to further intuitively understand and evaluate our model, in this section, we visualize the prediction of our STEP and the real data. 

\end{document}